\documentclass[lettersize,journal]{IEEEtran}
\usepackage{amsmath,amsfonts}
\usepackage{algorithmic}
\usepackage{algorithm}
\usepackage{array}
\usepackage[caption=false,font=normalsize,labelfont=sf,textfont=sf]{subfig}
\usepackage{textcomp}
\usepackage{stfloats}
\usepackage{url}
\usepackage{verbatim}
\usepackage{graphicx}
\usepackage{cite}
\usepackage{siunitx}
\usepackage{multicol}
\usepackage[%
    hidelinks=true]
    {hyperref}
\begin{document}

\title{A Dataset Fusion Algorithm for Generalised Anomaly Detection in Homogeneous Periodic Time Series Datasets}

\author{Ayman Elhalwagy and Tatiana Kalganova
\thanks{This work was sponsored in part by Voltvision}
}



\maketitle

\begin{abstract}
The generalisation of Neural Networks (NN) to multiple datasets is often overlooked in literature due to NNs typically being optimised for specific data sources. This becomes especially challenging in time-series-based multi-dataset models due to difficulties in fusing sequential data from different sensors and collection specifications. In a commercial environment, however, generalisation can effectively utilise available data and computational power, which is essential in the context of Green AI, the sustainable development of AI models. This paper introduces "Dataset Fusion," a novel dataset composition algorithm for fusing periodic signals from multiple homogeneous datasets into a single dataset while retaining unique features for generalised anomaly detection. The proposed approach, tested on a case study of 3-phase current data from 2 different homogeneous Induction Motor (IM) fault datasets using an unsupervised LSTMCaps NN, significantly outperforms conventional training approaches with an Average F1 score of 0.879 and effectively generalises across all datasets. The proposed approach was also tested with varying percentages of the training data, in line with the principles of Green AI. Results show that using only 6.25\% of the training data, translating to a 93.7\% reduction in computational power, results in a mere 4.04\% decrease in performance, demonstrating the advantages of the proposed approach in terms of both performance and computational efficiency. Moreover, the algorithm's effectiveness under non-ideal conditions highlights its potential for practical use in real-world applications.
\end{abstract}


\section{Introduction}
\IEEEPARstart{G}{eneralisation} is a measure of a Neural Network’s (NN) performance on data that it has not seen before but that is in the same class as the data that it has been trained on. The idea behind generalisation with Deep Learning (DL) is to transfer domain knowledge from data the NN has been trained on to unseen data in the same class, where the unseen data may contain conditions that slightly vary from the training data. This allows for a NN to be able to maintain performance across the dataset, and potentially transfer across multiple datasets with a similar distribution to the initial trained data. 
Various studies have been undertaken to understand the factors that affect the generalisation performance of a NN \cite{Jin2020}, and how to mitigate these factors to achieve the optimal level of performance \cite{Partridge1995}. The underlying concept is as follows: When training a NN on a data sample, the NN learns to represent the function between the input data and the output data through the adjustment of the weights and biases. If the distribution of the data sample used for training is not fully representative of the true distribution population, then the input-output function that is represented by this sample will inevitably vary from the function of the population. Extensive training on this sample will then result in a phenomenon known as overfitting \cite{Lawrence2000}, which refers to when the NN has accurately modelled the function represented by the training data but is not able to generalise to data in the same class due to the discrepancy between the functions represented by the sample and population.

Many works have been published with the aim of addressing overfitting and thus maximise the potential generalisation ability of a NN. Many works focus on architectural improvements to the NN to increase the robustness of the NN to overfitting through novel architectures such as Capsule Networks \cite{Elhalwagy2022}, whilst other research directions focus on the manipulation of the training procedure to limit overfitting with techniques such as Dropout \cite{Srivastava2014}, Early Stopping \cite{Caruana2001}, Pruning \cite{Wang2020} and adding noise to the weights and biases of the NN whilst training \cite{An1996}. 

There is much less focus on research concerning the effect of the composition and specification of the dataset on generalisation, especially regarding time series data. A common approach currently used include denoising the training sample to better align the distribution of the sample to the population and hence limit the level of overfitting \cite{Chen2021}\cite{Gholamreza2014}. Additionally, there is a consensus in ML research that increasing the volume of data through various means such as augmentation \cite{Shorten2019}\cite{Tanner1987} improves NN generalisation performance; whilst this has been empirically confirmed, more recent research has discovered that this improvement largely comes specific samples within the supplementary data, and a significant volume of this data is essentially redundant and does not contribute to a performance improvement \cite{Byerly2022a}\cite{Byerly2022b}.

\IEEEpubidadjcol
Furthermore, there is a gap in literature concerning the fusion of multiple time series datasets in a single training set to balance the probability distribution of the training sample so that it better aligns with the true distribution of the problem domain. This is largely due to a lack of necessity in an experimental environment since most ML research tends to optimise the solution for a single dataset source. However, from a commercial standpoint, this can have many benefits with regards to time and computational power saved, as well as an added benefit of reducing the data requirements for training. In addition to this, the dynamic shifting of the distribution of data is often a bottleneck to the performance of the NN; this is a prevalent issue that is encountered when a NN is deployed in a non-stationary environment, which is common with time series data. Some recent works have detected this shift \cite{Raza2013}, and mitigate the effect this has on the generalisation performance of the NN with both time series data \cite{Guo2016}\cite{Raza2014} and image data \cite{Cai2021}. However, the majority of empirical evaluations of NN approaches in literature are mostly conducted on an isolated sample of data, which, in many cases, is not representative of the dynamic shifting of the distribution temporally.

To address the identified gaps in the literature, we propose a novel algorithm, named Dataset Fusion. The proposed method merges multiple homogeneous, periodic time series datasets into a single unified dataset for training anomaly detection NN models. The fusion process is designed to accurately represent population distributions and increase robustness against potential data distribution shifts. Our primary objective in this study is to examine efficient generalisation approaches that can minimise training time and computational demands for neural networks when working with new homogeneous time series data sources. The contributions of this paper can be summarised as follows:

\begin{itemize}
\item{A novel dataset composition algorithm is proposed, referred to as \textit{Dataset Fusion}}
\item{The proposed approach is applied to a case study focused on motor current data, with a qualitative analysis conducted to assess the preservation of features from each individual dataset.}
\item{The generalisation performance of the proposed method is empirically evaluated in anomaly detection with the LSTMCaps neural network architecture from previous work \cite{Elhalwagy2022}, and compared to the performance when using conventional training approaches}
\item{The potential practical limitations of the proposed method in a real-world environment are discussed and assessed through further experimentation}
\end{itemize}

\section{Related Works}
This section investigates the current literature on different methods of addressing NN generalisation performance, as well as recent works using multiple datasets in different domains and tasks.

\subsection{NN based generalisation techniques}

\subsubsection{Dropout}
Dropout \cite{Srivastava2014} is a regularisation technique that can be used to prevent the neural network from overfitting on the training data. This is done by ignoring several randomly selected neurons, with the number of neurons dropped dependent on the “dropout rate” parameter, so they do not affect the output of the neural network on the forward pass. The idea behind dropout is to effectively train many subnets in your network so that your network acts as a sum of many smaller networks that can learn the representation of the data without the presence of the dropped-out neurons. This was found to improve the generalisation performance of the network by reducing data overfitting.

\subsubsection{Pruning}
Pruning is a process whereby a NN selectively removes trainable parameters based on an established criterion with the aim of maintaining the performance of the NN \cite{Han2015}. There are two main categories of pruning: Structured and Unstructured pruning. Unstructured pruning directly removes trainable parameters from the network, such as connections to neurons (weights). Structured pruning involves removing entire structures from the network such as neurons and filters. Structured pruning allows for a faster computational time in relation to unstructured pruning, as most frameworks existing for Machine Learning (ML) do not allow for the acceleration of sparse matrix calculations, therefore the NN will be able to reduce the number of calculations for the former but not the latter.

Various criterion has been established in literature for the pruning of NNs. One primitive but popular method is known as the weight magnitude criterion. The criterion dictates that the weights of the smallest magnitude are removed. The idea behind this is to remove all the weights that contribute the least to the function as they are less likely to impact the final prediction.

The most established framework for pruning is known as the train, prune and fine-tune method \cite{Han2015}. As the name suggests, the model is first trained, then iteratively pruned and fine-tuned based on the weight magnitude criterion. More recently however, an increasing number of works \cite{Qian2021} \cite{Urolagin2012} \cite{Wang2020} have evolved this framework with novel methodology that has allowed for further reduction of NN parameters and hence more efficient training and computation whilst maintaining similar performance.

\subsection{Data-based generalisation techniques}

Data-based generalisation techniques are largely overlooked in the field of deep learning in comparison to NN-based techniques. This is because the NN structure and learning optimisation algorithms are usually the reason for such weak performance, so improvements can largely be made by improving and optimising how the NN learns as opposed to what the NN is learning. However, it is still important to consider the training data as it can be a bottleneck for learning ability if not composed in the correct manner \cite{budach2022}.

An important aspect to consider is the difference in the data distribution between the data used to train the NN and the data that the NN will eventually be applied to. Often the data picked for training is not fully representative of the true distribution of the dataset, which creates a bottleneck to generalisation as the NN is not prepared for the distribution found in the overall population since it has been tuned to the distribution of the training sample. Shuffling the data before taking the training sample often helps with this to increase the likelihood of the sample representing the true distribution. Furthermore, shuffling during training also helps the NN weights to escape local minima and converge towards the global minimum of the function. Many works in deep learning falsely assume that the data distribution is static, which is largely incorrect as in practice data distributions are generally dynamic and tend to shift away from the test data distribution with real-world data; this phenomenon is known as distribution shift. This shift has been detected and quantified in recent works \cite{Lipton2018} \cite{Rabanser2019}, and accounted for with online adaptation to the shift \cite{Wu2021}, which allows the NN weights to adjust themselves to account for this distribution shift.

Data with excessive noise can inhibit the NNs ability to learn the true input-output function required by a specific application. Mathematically, this can be explained by the bias-variance decomposition \cite{neal2019modern}, specifically the variance component. The variance value refers to the change in prediction accuracy over different subsets of the data. A high variance value indicates that the NN has learnt the noise in the data, or in other words overfit on the training data. The most common method of reducing the variance is increasing the volume of the training data, which will naturally bring the distribution of the data closer to the required distribution that represents the overall dataset as opposed to just the subset of training data. As well as using real-world data, data augmentation has also proved to be an effective method of increasing data volume using data that is already accessible \cite{Shorten2019} \cite{Tanner1987}.  Another effective method that is commonly used is the denoising of data \cite{Fan2019}. This also contributes to the reduction of the variance: By removing the noise in the data, the training data subset will be cleaner and will more accurately represent the true distribution of the dataset. Whilst both methods are generally proven to work, there are still major issues with both methods that are still being addressed in literature, such as effectiveness in a real-world situation.

The financial and computational costs associated with increasing the volume of training data are not just substantial but also rapidly increasing \cite{Amodei2018}. This makes it increasingly difficult to train models without significant resources. Consequently, only well-funded entities with considerable resources can achieve a performance boost using this method, as their computational power and finances typically surpass those of other research entities and companies. This disparity creates a bottleneck for smaller entities, hindering their competitiveness in the field \cite{Yalalov2023}.

Denoising is still a very active field of research due to the many limitations of the currently proposed techniques. Since it is very difficult to determine which aspects of the data features are representative of the true distribution \cite{Fan2019}, it is very difficult to use denoising techniques such as filtering since features that are potentially important to NNs could be filtered out, limiting the potential performance of the NN. Furthermore, filtering techniques are largely contextual, so it is very difficult to develop filtering methods that work across multiple contexts to the same level of effectiveness.

Transfer learning \cite{Bozinovski2020} is also widely regarded in literature as a robust method of improving performance by utilising data from a different set but in the same domain as the target data. Whilst there have been many recent works regarding transfer learning \cite{Zhuang2021}, there is a gap in research concerning dataset fusion methods, an alternative to transfer learning that utilises multiple datasets in the same training phase, as opposed to multiple training phases to train a model more robust to the difference in data distributions between the training data and the data encountered during operation. This will be further explored in the present work.

\subsection{Multi-Dataset training}

Recently, researchers have identified the potential benefits of training generalisable NNs using multiple datasets to enhance performance and expand the capabilities of the NN models. Many of these approaches primarily focus on image-based applications \cite{Xian2018} \cite{Ranftl2022}. For example, Yao et al. \cite{Yao2020} introduced a novel framework for cross-dataset training, leveraging pre-existing labels from multiple datasets to create a single model capable of detecting the union of labelled features from all contributing datasets. This approach aims to maximise the utility of available labels for distinct classes in each dataset, thereby circumventing the time-consuming and resource-intensive process of labelling a single dataset with new classes that are already present in another dataset. Empirical evidence provided by the authors demonstrates the effectiveness of this approach when applied across multiple datasets, achieving comparable performance levels without sacrificing accuracy. In a related study, Zhou et al. \cite{Zhou2022} introduce a technique for training a unified object detection model on several extensive datasets by using dataset-specific training methods and losses, but maintaining a shared detection architecture with outputs specific to each dataset. This approach circumvents the necessity for manual taxonomy alignment, as it automatically combines the outputs of different datasets into a unified semantic taxonomy. The authors show that this multi-dataset detector achieves comparable performance to dataset-specific models in their respective domains while effectively generalising to unseen datasets without the need for fine-tuning. By utilising multiple training datasets, these approaches can lead to reduced resource requirements in terms of training data and improved performance. However, it is worth noting that they do not specifically address the aspect of reducing computational power during the training process.

In summary, analysing previous works on multi-dataset utilisation reveals clear benefits, such as reduced dataset labelling requirements and increased generalisation performance. However, there is a lack of exploration in time-series-based multi-dataset models due to challenges in fusing sequential data from different sensors and collection specifications. The current study aims to address these points by proposing a novel approach for effectively integrating time-series data from multiple sources. Furthermore, it is essential to consider computational efficiency, as training on multiple datasets may require additional computational resources. By addressing these points, this study aims to contribute valuable insights to the field of multi-dataset modelling and help pave the way for more robust and efficient approaches in various applications, particularly those involving time-series data.

\section{Dataset Fusion}
\label{section:Dataset Fusion}

\begin{figure*}[!t]
\centering
{\includegraphics[width=\linewidth]{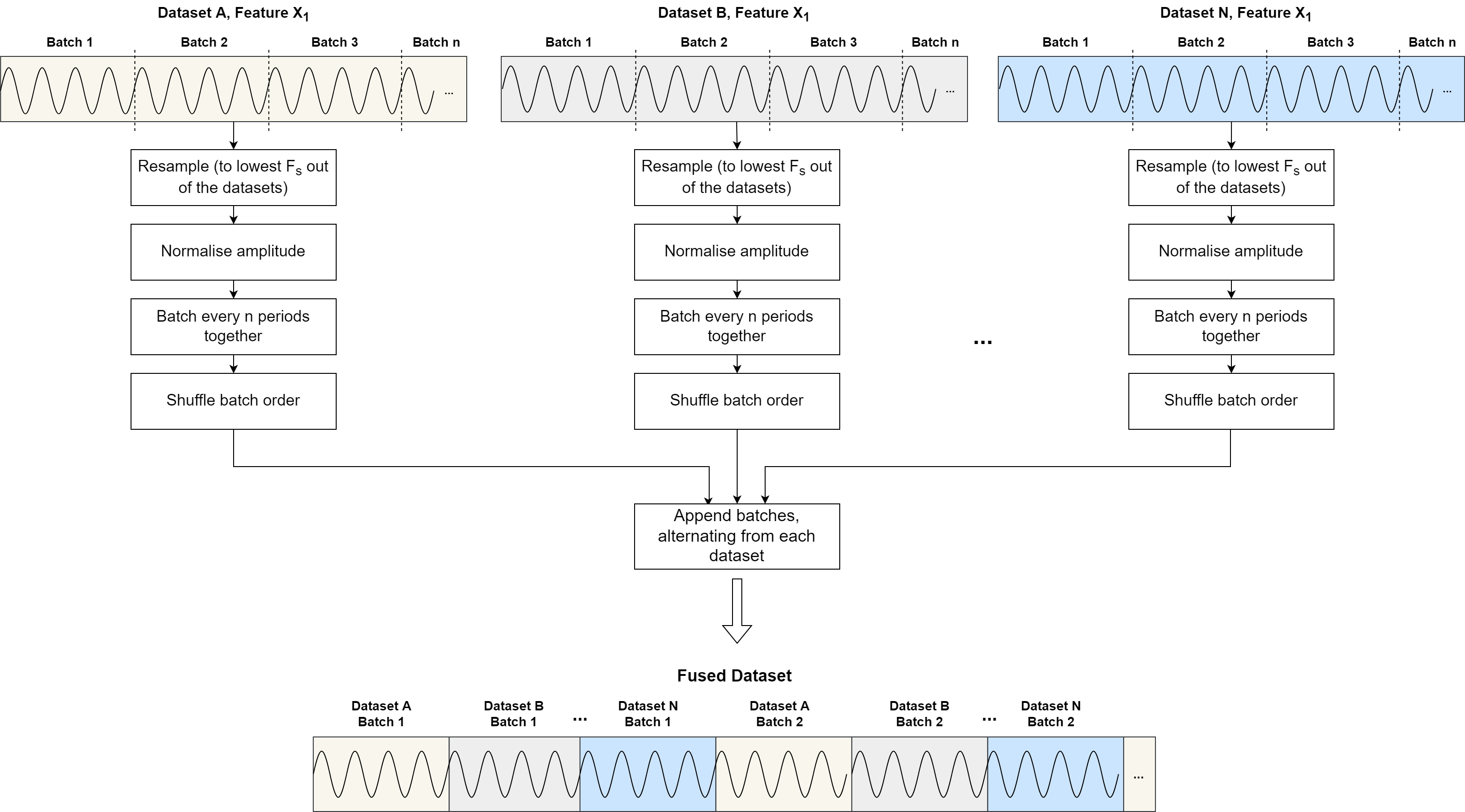}}%
\hfil
\caption{A summarised illustration of the Dataset Fusion algorithm}
\label{fig:dataset_fusion}
\end{figure*}

A summary of the dataset fusion process is depicted in Figure \ref{fig:dataset_fusion}.

The signal in each dataset is first down-sampled to the sampling frequency, $F_s$, of the dataset with the lowest sampling frequency. This step is essential to models with look-back such as Recurrent Neural Networks so that the same signal length is considered for every motor when training the model. Taking the example in Figure \ref{fig:dataset_fusion}, for a set of signals $\{A[n], B[n],… ,N[n]\}$ with sampling frequencies  $\{F_{s_{A}},F_{s_{B}},… ,F_{s_{N}}\}$, the target sampling frequency, $F_{s_{new}}$, is expressed in Equation \ref{eq:1}.

\begin{equation}
\label{eq:1}
    F_{s_{new}}=min\{F_{s_{A}},F_{s_{B}},… ,F_{s_{N}}\}
\end{equation}

The re-sampling is implemented using the Fourier method. This method was chosen over decimation due to the simplicity of implementation and with the assumption that the signals used are periodic in nature. To avoid aliasing and other artefacts, a low-pass windowed-sinc filter is first designed and applied to the signal, with a cutoff frequency based on the target Nyquist frequency. The Hann window, $hann(n)$, was employed in the filter design due to its desirable characteristics for resampling, such as reduced spectral leakage and smooth sidelobes. In the present work, 101 taps were used, giving a filter order of 100.  Equation \ref{eq:2} expresses the impulse response, $h[n]$, of this filter.

\begin{equation}
\label{eq:2}
h[n] = K \cdot \frac{\sin\left(2\pi f_c (n - M/2)\right)}{n - M/2} \cdot hann(n)
\end{equation}
\noindent where $K$ is the normalisation factor, $f_c$ is the cutoff frequency in Hz, $n$ is the discrete time index, and $M$ is the filter length or number of taps.

To implement the Fourier Method, the time series signal is first transformed into the frequency domain with a Discrete Time Fourier Transform (DTFT). For a discrete periodic sequence $x[n]$  its DTFT, $X[\omega]$ is expressed in Equation \ref{eq:3}.

\begin{equation}
\label{eq:3}
    X[\omega]=\sum_{n=-\infty}^{\infty} x[n]e^{-j\omega n}
\end{equation}

\noindent where $\omega=2 \pi f$ and $j=\sqrt{-1}$.

The spectrum $X[\omega]$ is then band-limited to the Nyquist frequency of the lowest $F_s$. The resulting spectra are then inverse-transformed back into the time domain. The down-sampling operation is expressed in Equation \ref{eq:4A} and the new number of samples, $N_{new}$, is calculated using Equation \ref{eq:4B}. Equation \ref{eq:5} shows the Inverse DTFT operation used to obtain the resampled time series signal.

\begin{subequations}\label{eq:4}
\begin{align}
    x[\omega]_{resampled} = \{x[\omega]:\omega<\frac{F_{s_{new}}}{2} \} \label{eq:4A}
\end{align}
\begin{align}
    N_{new} = \frac{N}{F_{s}}*F_{s_{new}} \label{eq:4B}
\end{align}
\end{subequations}

\begin{equation}
\label{eq:5}
    x[n]_{resampled}= \frac{1}{N_{new}} \sum_{\omega=0}^{N_{new}-1} X[n]e^\frac{j2 \pi\omega n}{N_{new}}
\end{equation}

Each dataset is then normalised using Z-score normalisation, to overcome varying motor currents. For the resampled sequence, $x[n]_{resampled}$, the normalised sequence, $x'[n]$ , is calculated using Equation \ref{eq:6}.
\begin{equation}
\label{eq:6}
    x'[n]=\frac{x[n]_{resampled}-\bar{x}_{resampled}}{\sigma_{x_{resampled}}}
\end{equation}
\noindent where $\bar{x}_{resampled}$ is the mean of the resampled sequence, and $\sigma_{x_{resampled}}$ is the standard deviation of the re sampled sequence.

To batch the periods together, a zero-crossing algorithm, configured to detect crossings from positive to negative, is employed to first identify a single period, and then concatenate n periods based on the user-defined parameter. Given that z-score normalisation is utilised, any periodic time-series data will exhibit sign changes, making the zero-crossing algorithm applicable. In cases where multiple features are present in the data, the zero-crossing algorithm calculates only the first feature as a reference for period batching, in order to maintain the temporal integrity of the data and preserve the spatial relationship between features. The impact of varying batch sizes on training performance differs across data types and problem domains; hence, it is recommended that this parameter be optimised alongside other training hyperparameters. For a discrete periodic signal $x[n]$ with assumed periodic sign changes, the set of indices for the positive-to-negative zero crossings, $c_{+ \, \rightarrow{\, -}}$, can be expressed as shown in Equation \ref{eq:7}.

\begin{equation}
\label{eq:7}
    c_{+ \, \rightarrow{\, -}} = \{\,  n \  | \ x[n-1] \,  > 0 \,  \geq x[n] \, \}
\end{equation}

For each dataset, the batch order is then shuffled randomly. This is done in order to mitigate the effect of distribution shift and prevent noise in one area of the signal from being prevalent in other areas of the signal. In other words, this step helps to reduce the variance of the NN prediction. A new signal is then constructed using the shuffled batches from each dataset by appending a batch from each dataset in an alternating fashion.

The full process of the Dataset Fusion algorithm is expressed in Algorithm \ref{alg:1}.

\begin{algorithm}[H]
\caption{Pseudocode of Dataset Fusion Algorithm}\label{alg:alg1}
\begin{algorithmic}
\STATE \textbf{Function:} Dataset\_Fusion($x,F_{s},P$)
\STATE \textbf{Input:} 
\STATE \hspace{0.5cm} A set of $s$ finite discrete periodic sets $x$ where $s>1$
\STATE \hspace{0.5cm} A set of sampling frequencies $F_{s}$ corresponding to $X$
\STATE \hspace{0.5cm} Number of periods batched $P$
\STATE \textbf{Output:} $ x_{fused} $
\STATE Determine $F_{s_{new}}$ using Equation \ref{eq:1}
\STATE \textbf{for} $x_{1} $ to $x_{s} $ \textbf{do}
\STATE \hspace{0.5cm} \textbf{if} $F_{s_{X_{s}}} \neq F_{s_{new}}$ \textbf{then}
\STATE \hspace{1cm} Apply filter shown in Equation \ref{eq:2}
\STATE \hspace{1cm} Calculate $X[\omega]$ using Equation \ref{eq:3}
\STATE \hspace{1cm} Calculate $X_{[\omega]_{resampled}}$ using Equation \ref{eq:4A}
\STATE \hspace{1cm} Calculate $N_{new}$ using Equation \ref{eq:4B}
\STATE \hspace{1cm} Calculate $x[n]_{resampled}$ using Equation \ref{eq:5}
\STATE \hspace{0.5cm} Calculate $x'$ using Equation \ref{eq:6}
\STATE \hspace{0.5cm} Calculate $c_{+ \, \rightarrow{\, -}}$ using Equation \ref{eq:7}
\STATE \hspace{0.5cm} Calculate $x'_{batched}$ through grouping $P$ periods by 
\STATE \hspace{0.5cm} slicing $x'$ at the values where $c_{+ \, \rightarrow{\, -}}[n] \% P = 0$
\STATE \hspace{0.5cm} Shuffle $x'_{batched}$
\STATE $x_{fused} = \{x_{1_{batched}}'[0] +\!\!\!+ ... +\!\!\!+ x_{s_{batched}}'[0] +\!\!\!+ x_{1_{batched}}'[1]$
\STATE $+\!\!\!+ ...+\!\!\!+ x_{s_{batched}}'[1]+\!\!\!+ ...\}$
\STATE \textbf{return} $x_{fused}$
\end{algorithmic}
\label{alg:1}
\end{algorithm}
\noindent where $\%$ represents the modulo operator and $+\!\!\!+$ represents concatenation.

\subsection{Computational Complexity}

The computational complexity of the Dataset Fusion algorithm, represented in Big O notation, can be determined by breaking down the steps of the algorithm when practically applied. The breakdown of the complexity of each stage is provided in Table \ref{tab:1}.

\begin{table}[!t]
\caption{Algorithm Complexity for Dataset Fusion Algorithm, for $n$ Datasets with $m$ length}
\label{tab:1}
\begin{tabular}{ll}
\hline
Algorithm step                & Big O Complexity                       \\ \hline
Filtering and Resampling      & $O(nm(1 + log(m)))$                \\
Normalisation                 & $O(nm)$                               \\
Period Batching               & $O(nm)$                               \\
Chaining and Stacking batches & $O(nm)$                               \\
Total                         & $O(nm (1 + log(m))) + 3 O(nm)$ \\ \hline
\end{tabular}
\end{table}

As Table \ref{tab:1} shows, the filtering and resampling step has a complexity of $O(nm(1 + log(m)))$, since it involves applying a finite impulse response (FIR) filter with a complexity of O(m) and performing resampling using the Fast Fourier Transform (FFT) with a complexity of $O(m * log(m))$ for each of the n datasets. The Normalisation step scales each dataset using Z-score normalisation with a complexity of $O(m)$ for each dataset, resulting in a total complexity of $O(n * m)$. The Period Batching step identifies zero-crossings and creates period batches with a complexity of $O(m)$ for each dataset, resulting in a total complexity of $O(n * m)$. Finally, the Chaining and Stacking batches step involves filtering, chaining, and stacking the period batches with a total complexity of $O(n * m)$. The overall complexity of the Dataset Fusion algorithm is the sum of the complexities of these steps, which is $O(n * m * (1 + log(m))) + 3 * O(n * m)$, with the dominating term being $O(n * m * log(m))$ due to its faster growth as the input size (n and m) increases.

The logarithmic factor in the dominating term, $O(n * m * log(m))$, makes the Dataset Fusion algorithm scale well with increasing input size. This is because logarithmic growth is slow growth, ensuring that the algorithm remains efficient even as the number and size of the datasets (n and m) increase. Additionally, since the complexity is dependent on both the number of datasets (n) and the length of the datasets (m), the algorithm can efficiently handle varying dataset sizes and compositions. This scalability makes the Dataset Fusion algorithm a versatile algorithm and suitable for processing large and diverse datasets, which is essential in the context of real-world applications where data size and complexity are constantly evolving.

\subsection{Requirements for application}

Whilst the proposed algorithm is domain-independent, there are requirements regarding the data that must be met for the proposed method to be applicable. These requirements, as well as the reasoning, are detailed in the following sections.

\subsubsection{Homogeneous Datasets}

Although the methodology can be used in varying problem domains, the dataset fusion algorithm can only fuse homogeneous data, since the aim of the algorithm is to capture the data distribution of a problem domain as a whole in order to mitigate overfitting on a specific dataset. Generalisation to multiple problem domains is not in the scope of this algorithm.

\subsubsection{Data periodicity}

As explained in section \ref{section:Dataset Fusion}, The algorithm relies on the fact that the data is periodic, due to the resampling method used, the zero crossing method, and to be able to create a coherent and usable sequential fused time series dataset.

\subsubsection{Time Domain Data}

The proposed approach will only be applicable in the time domain representation of the datasets, as it relies on the sequential nature of the data to fuse it together in a meaningful way

If the datasets being fused meet the requirements detailed above, then there is feasibility in applying the proposed method. Some examples of where the proposed method may be feasible are daily temperatures in a region, electrical power data and vibration data.

\subsection{Proposed Benefits of Dataset Fusion}

The Dataset Fusion algorithm seeks to eliminate the necessity of multiple NNs for a single problem domain. This approach theoretically enables the development of an NN that can adapt to unseen data from the same domain, even if originating from different data sources. Moreover, the Dataset Fusion algorithm aims to reduce data requirements from individual sources, as achieving ideal data collection conditions from each source can often prove to be challenging. Potential issues with collected data, such as data corruption, sensor faults, or insufficient data volume, among other data collection complications, further emphasise this need. The present study will experimentally investigate the proposed benefits of the Dataset Fusion algorithm.

\section{Case Study: Dataset Fusion for 3-phase motor current data}

This section will explore the feasibility of applying the proposed method with a case study on motor current signals. The aim of the case study is to empirically test and validate the effectiveness of our proposed method. The datasets used will first be introduced and the feasibility of the proposed method will be confirmed. The dataset fusion algorithm will then be applied, and the resulting signal will be compared and analysed to the original signals.

\subsection{Dataset Introduction}

For the present case study, two homogeneous open-source datasets \cite{Maciejewski2020} \cite{Cunha2020} will be used to confirm the feasibility of the Dataset Fusion methodology. Specifications of the datasets used are detailed in Table \ref{tab:2}. Both datasets are composed of three-phase motor current signals, however, one dataset, which will be referred to as Dataset A, contains fault data for an inter-turn short circuit fault, and the other dataset, which will be referred to as Dataset B, contains current signals for a broken rotor bar fault. 

\begin{table*}[!t]
\centering
\caption{Specification for Motor Datasets used in the case study}
\label{tab:2}
\resizebox{\textwidth}{!}{%
\begin{tabular}{lrrrrr}
\hline
Dataset Name                                                    & \begin{tabular}[c]{@{}l@{}}Data volume per file\\ (samples, features)\end{tabular} & \begin{tabular}[c]{@{}r@{}}Faulty Data\\ Files\end{tabular} & \begin{tabular}[c]{@{}r@{}}Healthy Data\\ Files\end{tabular} & \begin{tabular}[c]{@{}r@{}}Sampling \\ Frequency (Hz)\end{tabular} & \begin{tabular}[c]{@{}l@{}}Duration\\ per file(s)\end{tabular} \\ \hline
Dataset A: Inter-turn short circuit fault dataset (Cunha, 2021) & (100,000, 3)                                                                        & 2264                                                        & 353                                                          & 10,000                                                             & 10                                                             \\
Dataset B: Broken Rotor Bar Dataset (Maciejewski, 2020)         & (1,001,000, 3)                                                                       & 320                                                         & 80                                                           & 55,611                                                              & 18                                                             \\ \hline
\end{tabular}%
}
\end{table*}

\subsubsection{Dataset A}

Dataset A \cite{Cunha2020} contains files from a motor running at a variable operating frequency $F_o$, ranging from 30Hz to 60Hz with 5Hz increments. The motor has the following specifications: 4 poles, 1HP mechanical power, 220V supply and 3A rated current. The authors simulated both high-impedance and low-impedance short circuits, for different levels of fault severity. For the purpose of this case study, only the files captured at $F_o=60Hz$ were used to meet the limitation of Dataset Fusion of only being applicable to homogeneous data. The new dataset structure is depicted in Table \ref{tab:3}.

\subsubsection{Dataset B}

The motor used to capture Dataset B \cite{Maciejewski2020} is a squirrel cage AC motor, running at a constant $F_o=60HZ$ and has similar specifications to the motor used to capture Dataset A. The breakdown of the dataset is shown in Table \ref{tab:4}. At the beginning of each file, for roughly the first 4 seconds, a transient signal representing the motor startup was also recorded. For the purpose of the case study, and for compatibility with the proposed algorithm, the transient subset of the signal, the first 200,000 samples, was discarded from each file, so that only the steady state of the motor remained. This left 801,000 samples left in each file, representing an approximate 20\% redundancy of data.

\begin{table}[]
\centering
\caption{Breakdown of Dataset A 60Hz files}
\label{tab:3}
\begin{tabular}{lll}
\hline
Motor State                                                                           & Number of files & Samples per feature\\ \hline
Healthy                                                                               & 48              & 4,800,000                                                                        \\
\begin{tabular}[c]{@{}l@{}}High Impedance 1\\ (1.41\% of stator winding)\end{tabular} & 53              & 5,300,000                                                                        \\
\begin{tabular}[c]{@{}l@{}}High Impedance 2\\ (4.81\% of stator winding)\end{tabular} & 52              & 5,200,000                                                                        \\
\begin{tabular}[c]{@{}l@{}}High Impedance 3\\ (9.26\% of stator winding)\end{tabular} & 48              & 4,800,000                                                                        \\
\begin{tabular}[c]{@{}l@{}}Low Impedance 1\\ (1.41\% of stator winding)\end{tabular}  & 55              & 5,500,000                                                                        \\
\begin{tabular}[c]{@{}l@{}}Low Impedance 2\\ (4.81\% of stator winding)\end{tabular}  & 54              & 5,400,000                                                                        \\
\begin{tabular}[c]{@{}l@{}}Low Impedance 3\\ (9.26\% of stator winding)\end{tabular}  & 60              & 6,000,000                                                                        \\ \hline
\end{tabular}
\end{table}

\begin{table}[!t]
\centering
\caption{Breakdown of Dataset B}
\label{tab:4}
\begin{tabular}{lll}
\hline
Motor state   & Number of files & Samples per feature \\ \hline
Healthy       & 80              & 1,001,000           \\
1 Broken Bar  & 80              & 1,001,000           \\
2 Broken Bars & 80              & 1,001,000           \\
3 Broken Bars & 80              & 1,001,000           \\
4 Broken Bars & 80              & 1,001,000           \\ \hline
\end{tabular}
\end{table}

\subsection{Application of Dataset Fusion and Analysis}

The Dataset Fusion algorithm was used to fuse the healthy files from Datasets A and B into a single, fused dataset. First, all healthy files were extracted from each dataset and concatenated into a large signal. The files in Dataset B were first sliced to remove the motor startup signature, then resampled to 10,000Hz, the same $F_s$ as Dataset A. Each Dataset was split into batches of 4 periods and then concatenated alternating between each Dataset to create the final fused dataset.

\begin{figure*}[!t]
\centering
\setlength{\tabcolsep}{0pt}
\begin{tabular}{ccc}
\multicolumn{3}{c}{\textbf{Time Series Signals}} \\
\subfloat[]{\includegraphics[width=2.5in]{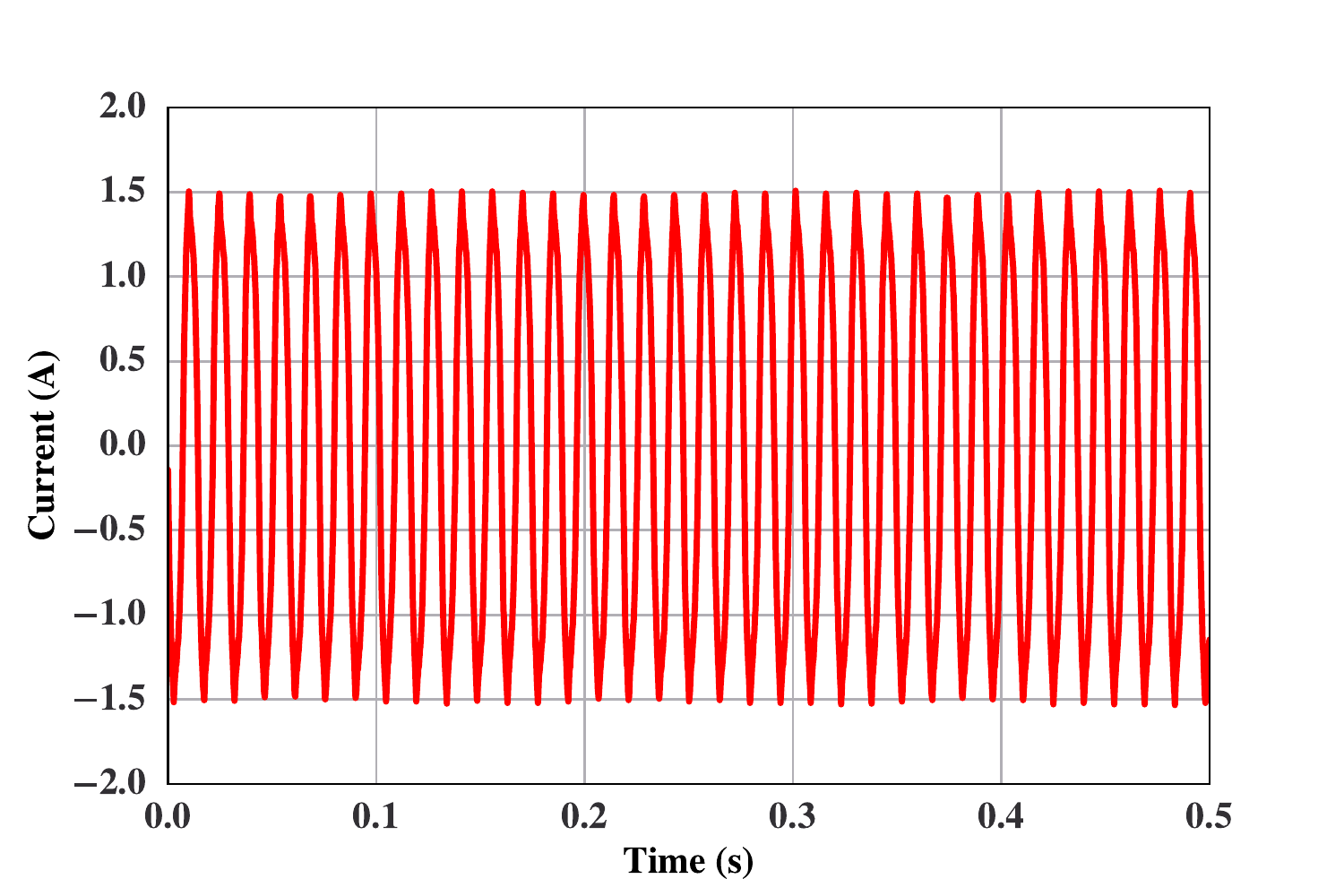}} &
\subfloat[]{\includegraphics[width=2.5in]{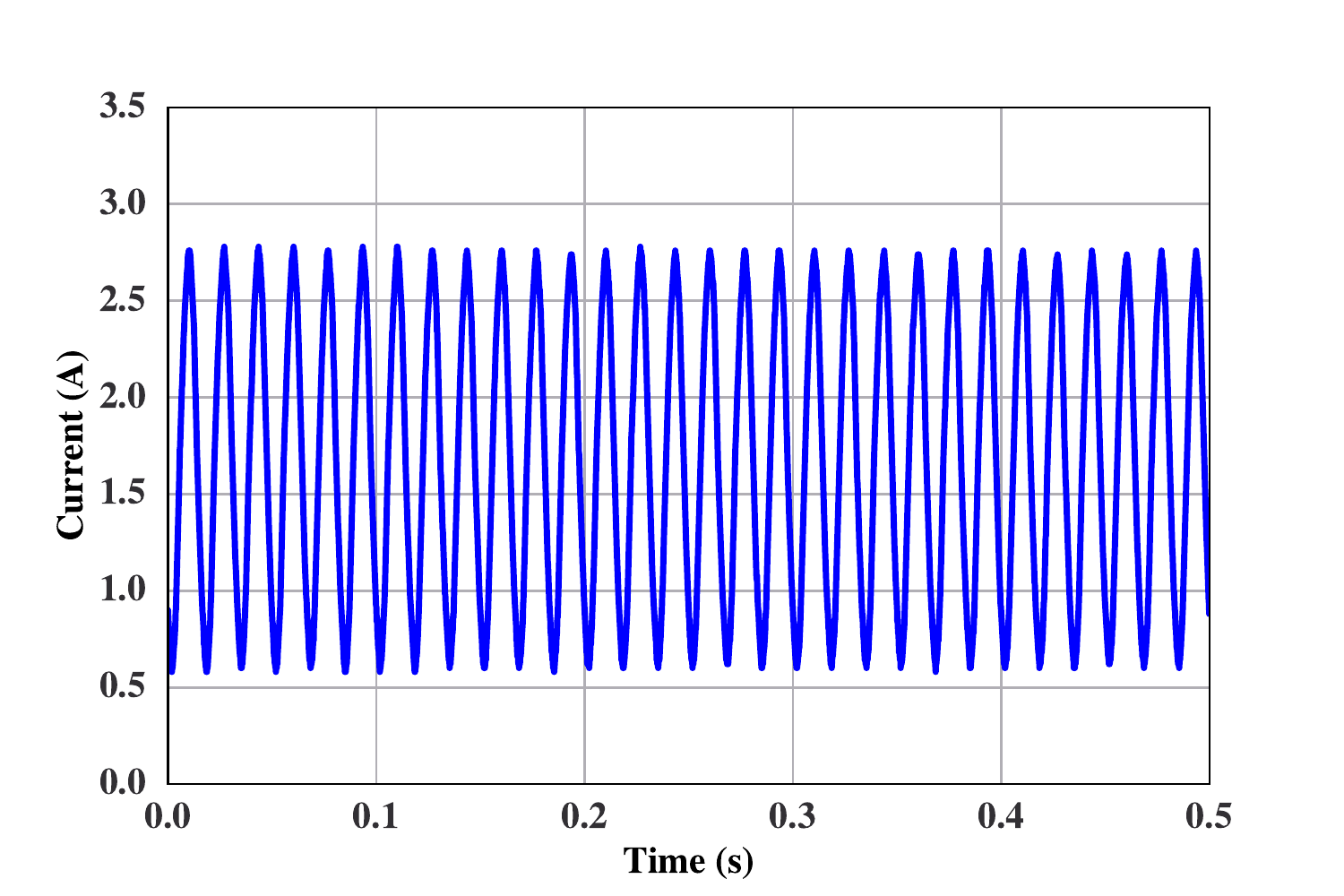}} &
\subfloat[]{\includegraphics[width=2.5in]{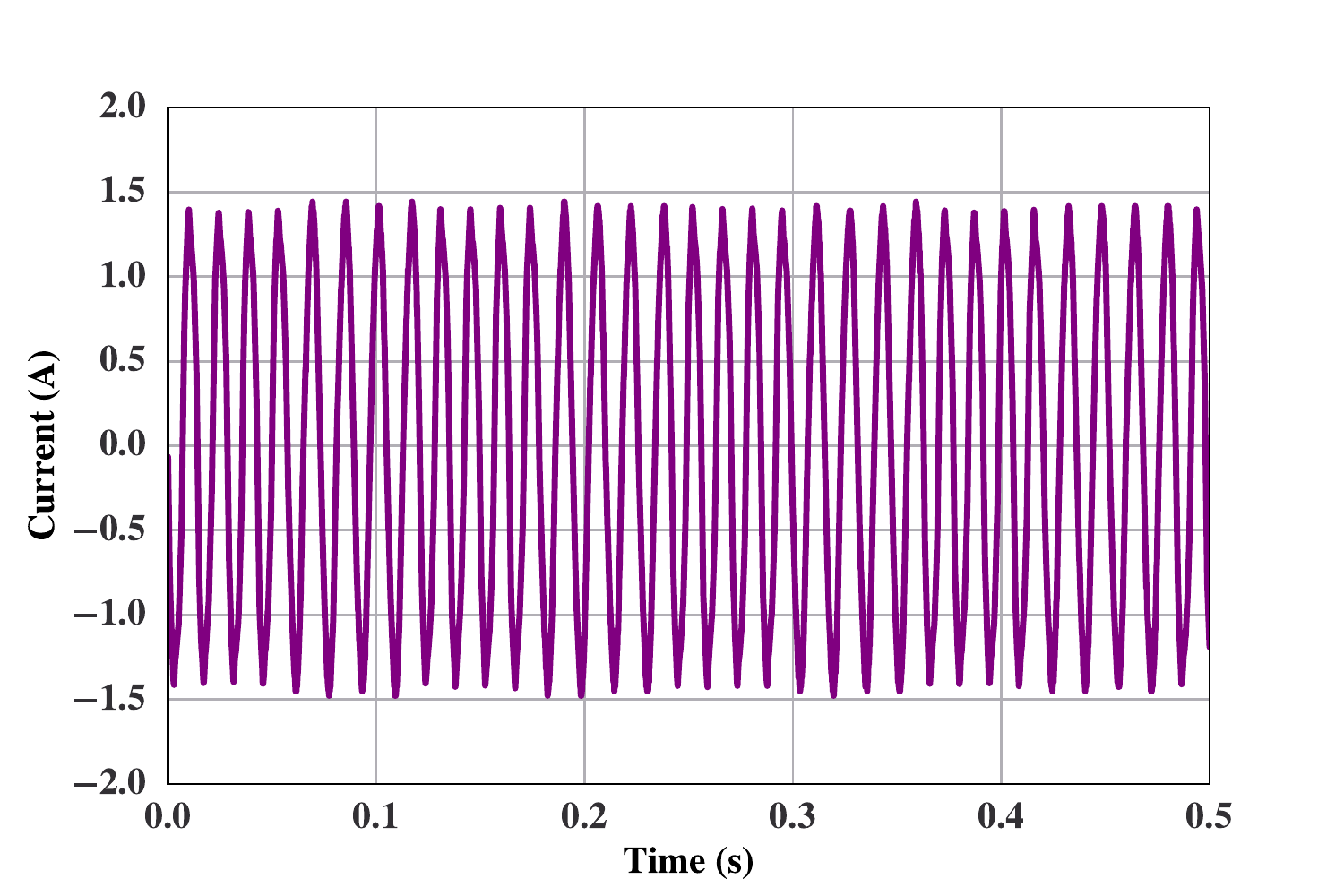}} \\
\multicolumn{3}{c}{\textbf{Probability Distribution Functions}} \\
\subfloat[]{\includegraphics[width=2.5in]{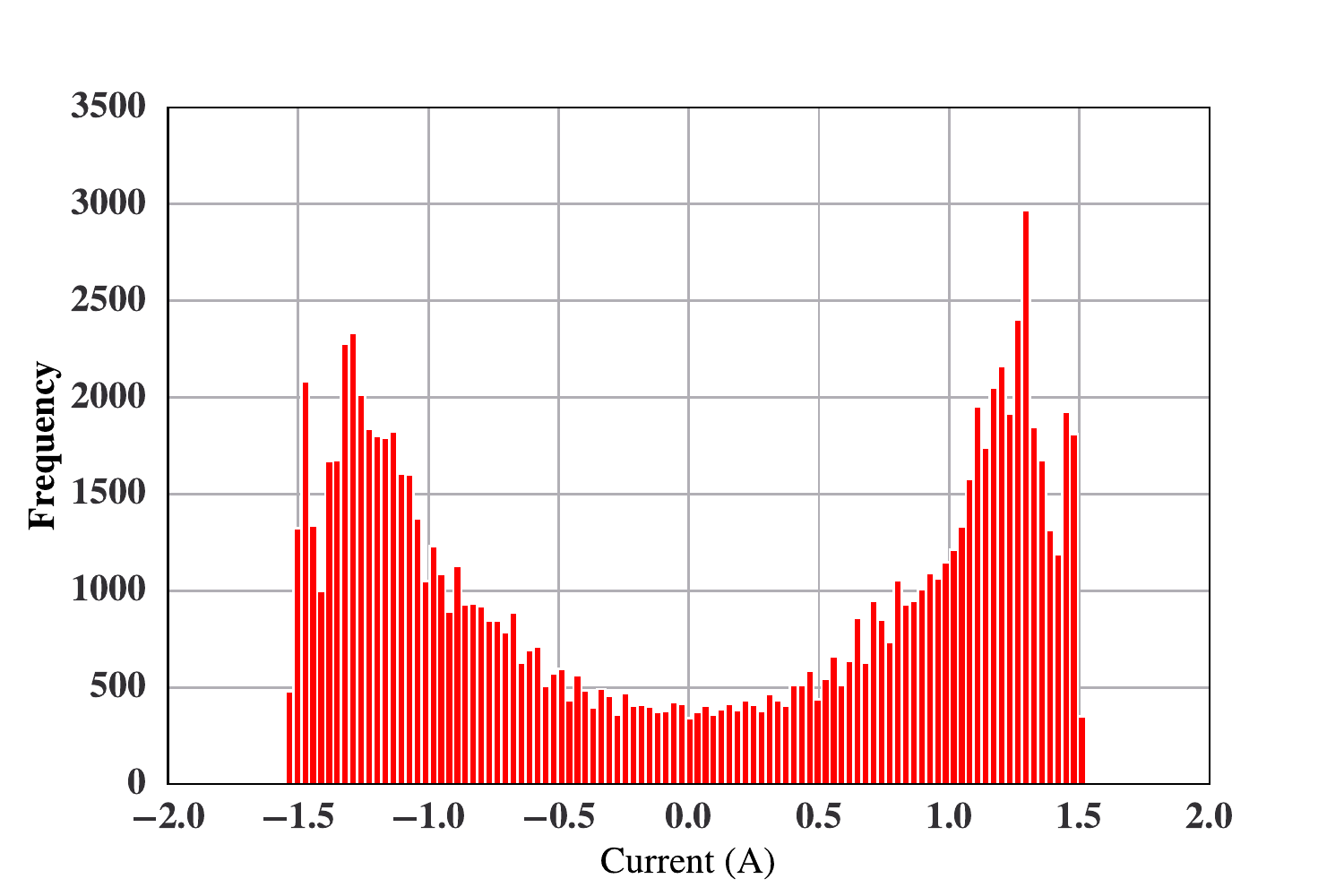}} &
\subfloat[]{\includegraphics[width=2.5in]{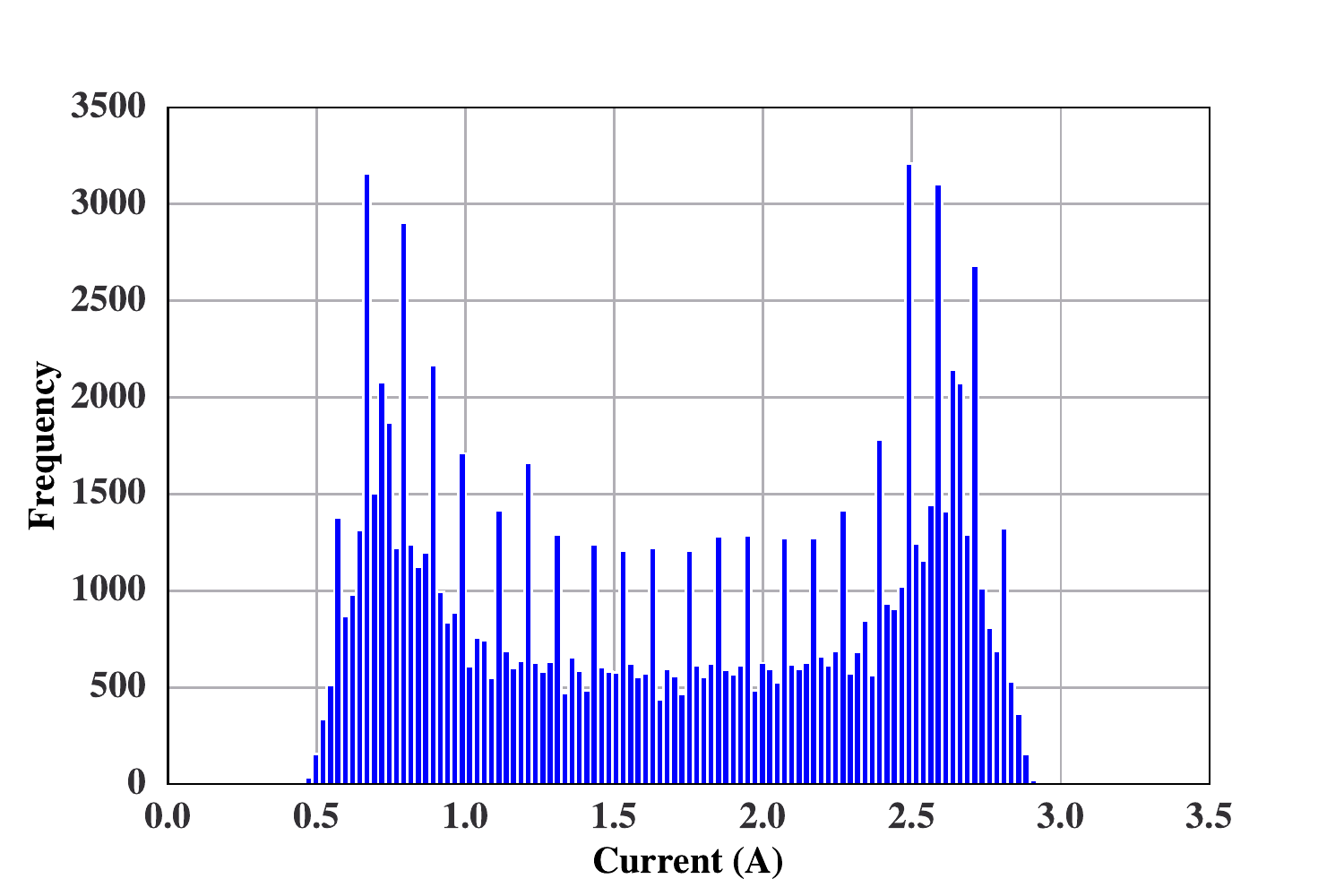}} &
\subfloat[]{\includegraphics[width=2.5in]{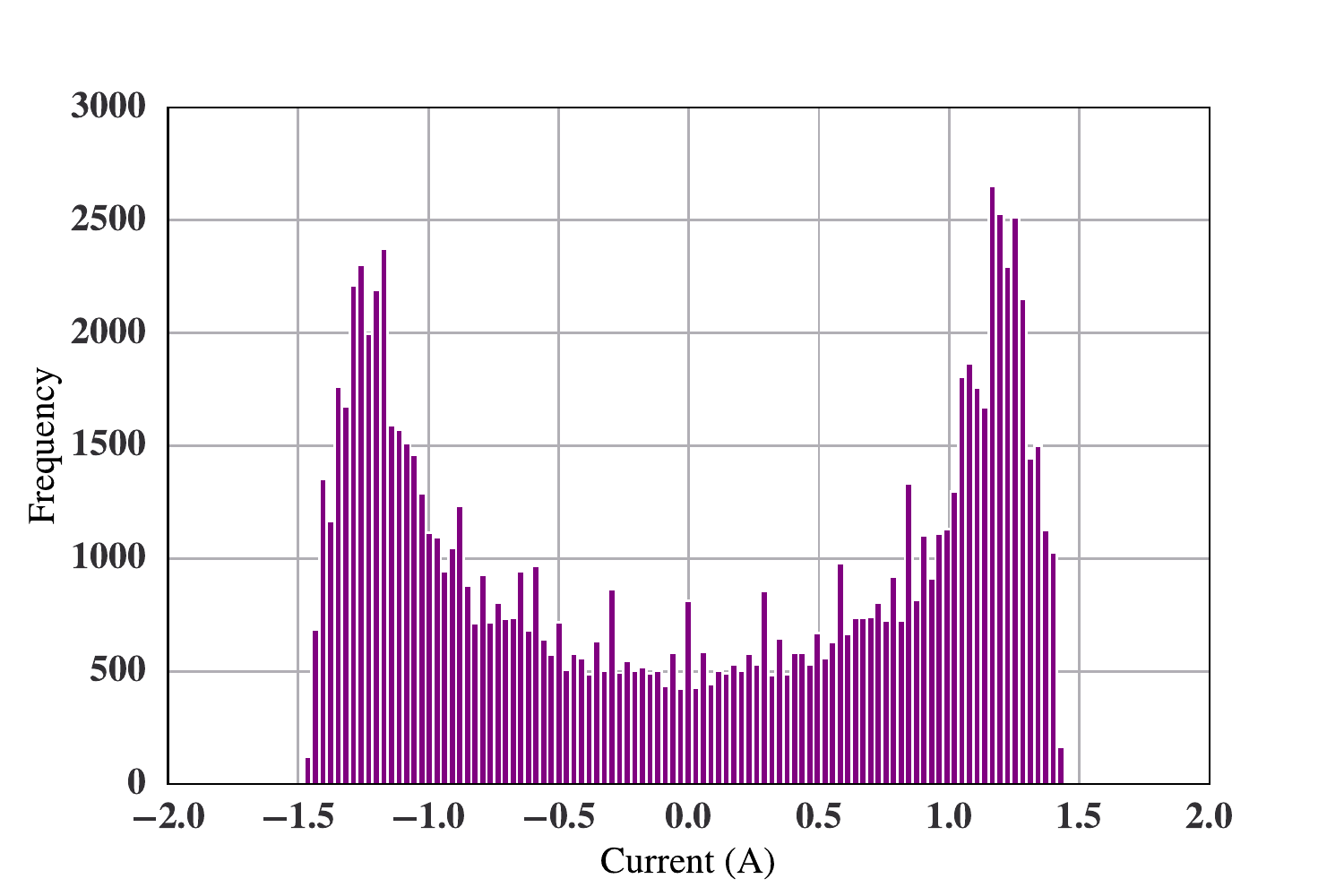}} \\
\multicolumn{3}{c}{\textbf{Fourier Transforms}} \\
\subfloat[]{\includegraphics[width=2.5in]{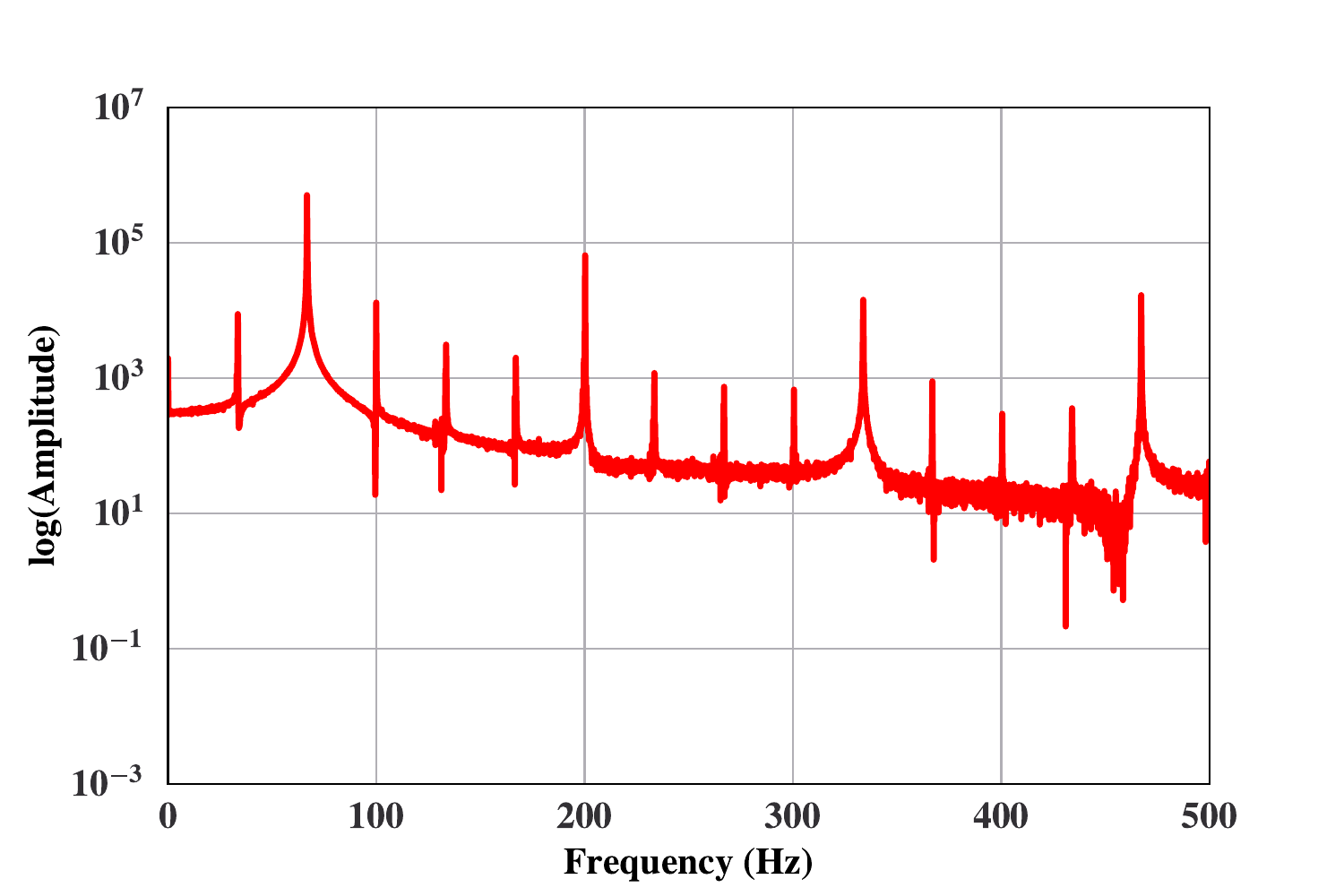}} &
\subfloat[]{\includegraphics[width=2.5in]{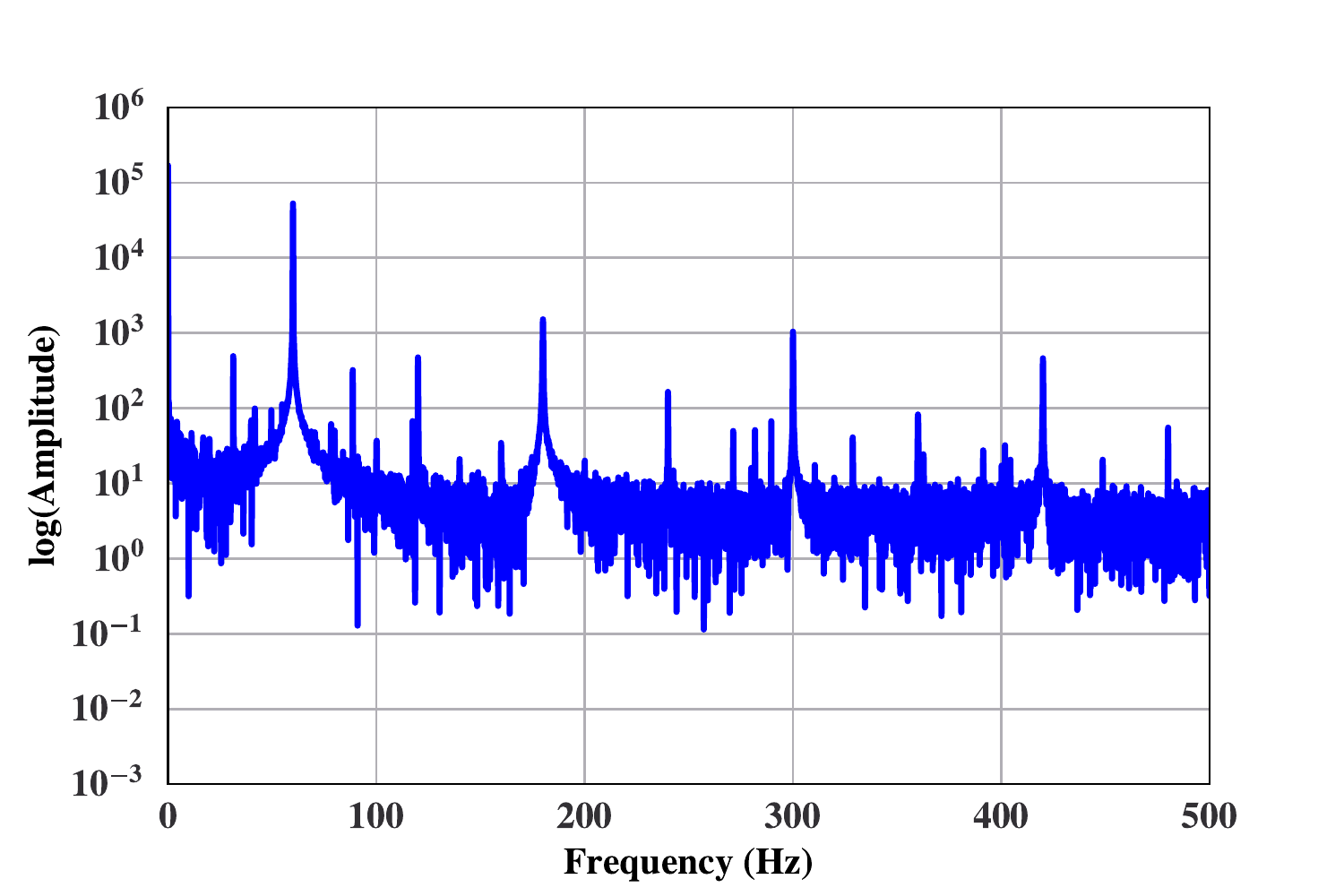}} &
\subfloat[]{\includegraphics[width=2.5in]{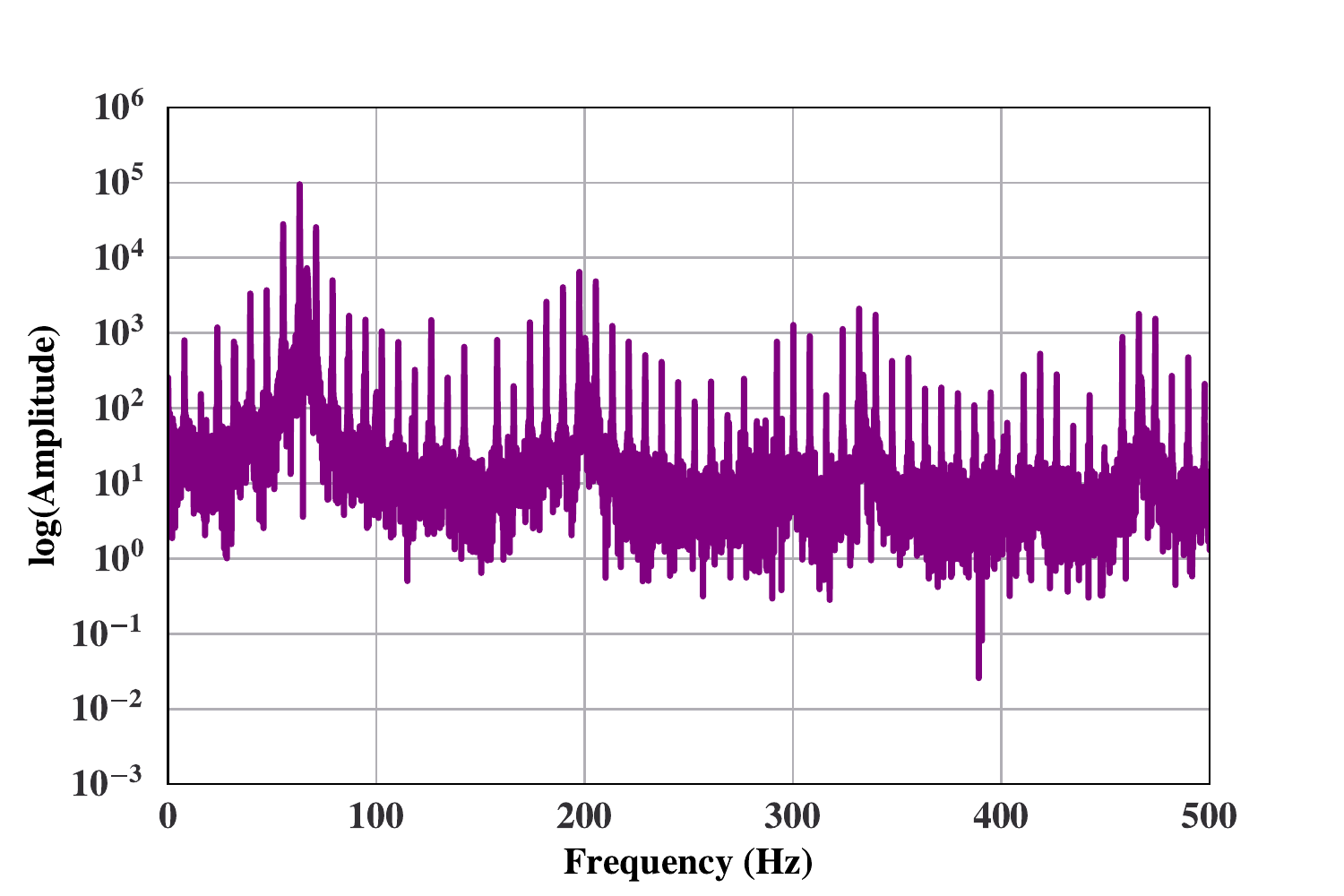}} \\
\end{tabular}
\caption{Time series signal from Dataset A (a), B (b) and fused (c), Probability Distribution Function from Dataset A (d), B (e) and fused (f), and Fourier Transform from Dataset A (g), B (h) and fused (i) healthy files}
\label{fig:2}
\end{figure*}

For each dataset, an initial analysis was conducted in order to understand the data and enable a more accurate interpretation of the fused dataset experimental results. The time series signal, a Probability Distribution Function (PDF) and FT representations from 0-500Hz were generated for a single phase from a healthy file from each dataset. The plots are illustrated in Figure \ref{fig:2}.

\begin{figure}[!t]
\centering
\includegraphics[width=2.5in]{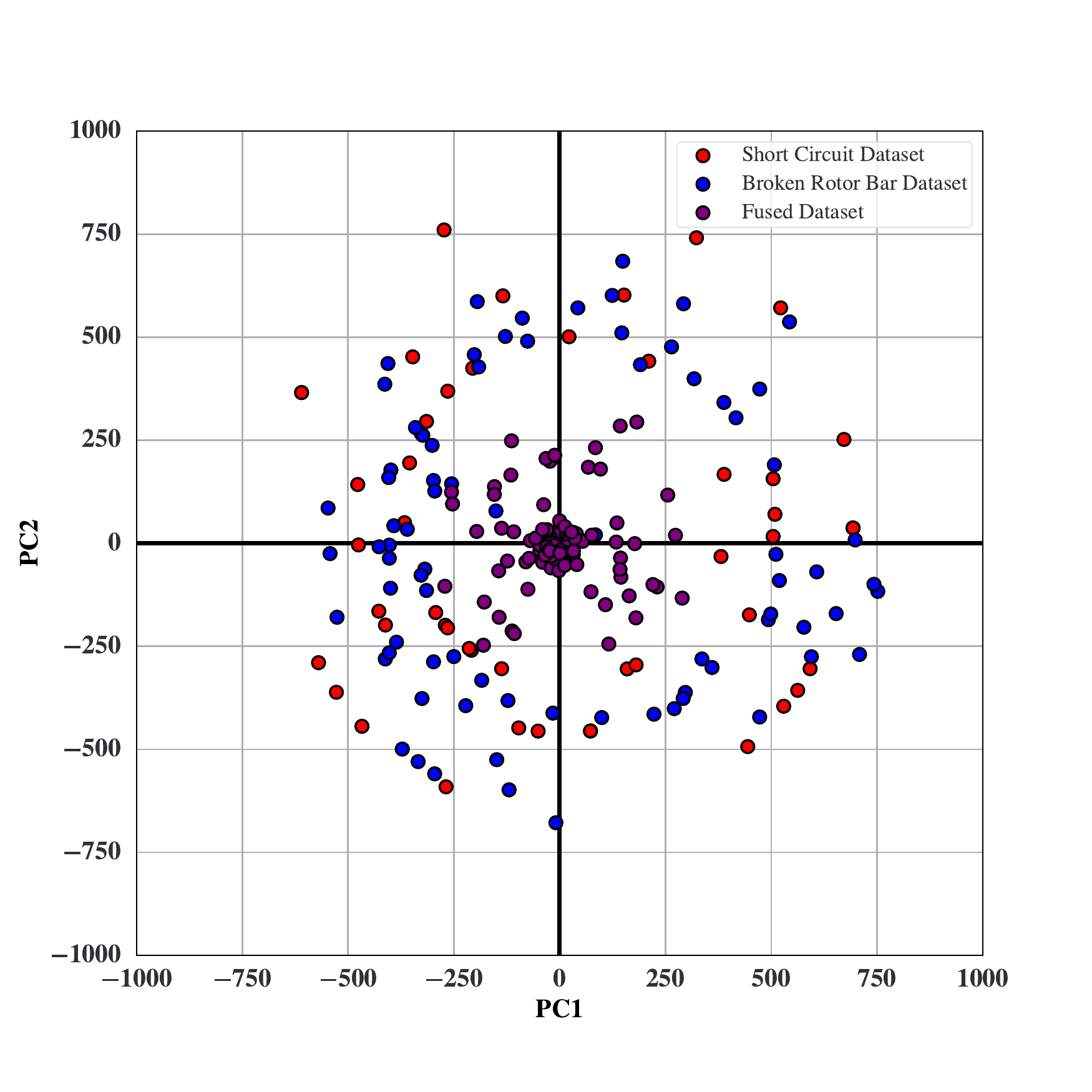}
\caption{Principal Component Analysis of Dataset A, B and the fused dataset}
\label{fig:3}
\end{figure}

A Principal Component Analysis (PCA) was also performed on the healthy data from each dataset as well as the fused data to gain comprehensive insights into the data, wherein the resulting axes are linear combinations of the original variables, defined by the eigenvectors and eigenvalues. This method allows for the identification of the most significant patterns to increase the interpretability of the proposed method. All datasets were uniformly downsampled to 10,000Hz, which corresponds to the minimum sampling frequency in Dataset A. Subsequently, the samples were partitioned into groups of 100,000, aligning with the smallest sample size per file in Dataset A. The data's three features, representing the three phases, were flattened into a single axis before being subjected to the PCA algorithm. The visualisation of the first two Principal Components in a 2D scatter plot can be found in Figure \ref{fig:3}.

It is clear to see from Figure \ref{fig:2}(d) and Figure \ref{fig:2}(e), as well as Figure \ref{fig:2}(g) and Figure \ref{fig:2}(h) that Dataset B contains a considerable amount of noise in comparison to Dataset A. In addition to this, the frequency spectra of Dataset A show more pronounced harmonics in comparison to Dataset B. Although this may not be as evident from the time series signal plot, a NN will most likely pick up these differences in noise, and thus a NN trained on a single dataset, especially in the case of Dataset B, will struggle to distinguish files from Dataset A with fault signals as anomalous. This hypothesis will be further discussed in the results and discussion.

The time series signal of the fused data looks similar visually in comparison to the datasets, albeit in a different input space due to normalisation. From the PDF and FFT representations shown in Figure \ref{fig:2}(f) and Figure \ref{fig:2}(I) however, there are subtle indications of features present in both Dataset A and Dataset B. For instance, the overall shape of the frequency spectra follows Dataset A, however, there is noise clearly present in the spectra, a significant feature of Dataset B. Furthermore, the harmonic peaks in the fused frequency spectra contain the same characteristics of both datasets, which is interesting to note as this representation would still be considered a healthy signal. Future work will investigate the use of a fused Dataset in the frequency domain to train a classifier NN. However, the scope of this study is to validate the use of the time series representation to train a generalised time series anomaly detector with reduced data requirements. 

Upon examining the PCA plot depicted in Figure \ref{fig:3}, it is clear that the healthy data from both datasets exhibit comparable traits and patterns. Interestingly, the fused dataset forms a cluster around the origin, positioning itself at the center of the two datasets. This central location of the fused dataset within the circular arrangement of points from the two datasets signifies that it effectively captures the salient features of both datasets. By doing so, the fused dataset aids in bringing the training data closer to the population distribution of the problem domain, thereby enhancing the robustness and generalisability of the model derived from this data. Further evidence of this will be given in the experimental results.

It is important to note that simply concatenating two healthy files from each Dataset will produce a similar outcome to the representations shown in Figure \ref{fig:2}. However, the purpose of this analysis is to show that the Dataset algorithm will still preserve the individual features of each representation in a new signal and still be usable for a data-driven approach. The PCA plot, as displayed in Figure \ref{fig:3}, provides more compelling evidence of the impact of the Dataset Fusion technique on the combined dataset. Subsequent experimental results on anomaly detection, utilising the various datasets, will further explore the implications of employing a fused dataset for training an anomaly detection model.

\subsection{Experimental Design} \label{Experimental Design}

The aim of the experimentation presented in this study is to observe the effectiveness of the Dataset Fusion with training an anomaly detector NN in comparison to commonly used training methods. The following training methods will be compared for all of the following experiments:

\begin{itemize}
    \item {\textbf{Traditional Approach}: Training on a single dataset}
    \item {\textbf{Transfer Learning}: First training phase on 1 dataset, second training phase on another dataset}
    \item {\textbf{Mixed Dataset}: Single training phase on all healthy files from each Dataset}
    \item {\textbf{Dataset Fusion}: Single training phase on fused healthy data consisting of all datasets}
    \item {\textbf{Dataset Fusion with Transfer Learning}: First training phase on fused healthy data consisting of all datasets, second training phase on single dataset}
\end{itemize}
Since each training method has multiple variants, for example, Transfer Learning with the order in which the datasets are used, Table \ref{tab:5} provides a full breakdown of the different variants, as well as the labels that will be used in the experimental results tables.

\begin{table}[]
\centering
\caption{Experiment variants and corresponding key for results tables}
\label{tab:5}
\begin{tabular}{ll}
\hline
Experiment                                     & Key       \\ \hline
Traditional Approach -  Dataset A              & T - Dataset A        \\
Traditional Approach - Dataset B               & T - Dataset B        \\
Transfer Learning - Dataset A to Dataset B     & TL - Dataset A to B \\
Transfer Learning - Dataset B to Dataset A     & TL - Dataset B to A \\
Mixed Dataset                                  & MD        \\
Dataset Fusion                                 & DF        \\
Transfer Learning - Fused Dataset to Dataset A & TL - DF to Dataset A \\
Transfer Learning - Fused Dataset to Dataset B & TL - DF to Dataset B \\ \hline
\end{tabular}
\end{table}

The same workstation was used to conduct all experimentation in order to maximise experimental rigour. The specifications of this workstation are given in Table \ref{tab:6}, for the purpose of experiment reproducibility.

\begin{table}[!t]
\centering
\caption{Specifications for workstation used for experimentation}
\label{tab:6}
\begin{tabular}{ll}
\hline
Component        & Specification                                          \\ \hline
Operating System & Windows 10 Version 21H2                                \\
CPU              & AMD Ryzen Threadripper 2990WX 32-Core 3.5GHz           \\
RAM              & 64GB                                                   \\
GPU              & NVIDIA RTX A6000 48GB VRAM                             \\ \hline
\end{tabular}
\end{table}

Each experiment iteration was repeated 10 times for experimental rigour. The outcome of each experiment is validated for statistical significance using an Analysis Of Variance (ANOVA), to ensure the reproducibility of the results presented. The ANOVA is generated using custom functions on Python 3.9, and the numpy \cite{harris2020array} and pandas \cite{reback2020pandas} libraries, with versions 1.22.0 and 1.3.5 respectively. 

\subsubsection{Neural Network Model}

The multi-channel LSTMCaps autoencoder NN developed in previous work will be used as the anomaly detection models for the following experiments. Further details regarding the architecture are given in \cite{Elhalwagy2022}. For the following experimentation, three input branches were used to accommodate the three features present in the dataset: the three-phase motor current signals. The model hyperparameters were optimised for anomaly detection using iterative sweeps. Table \ref{tab:7} details the optimal hyperparameters for this task, which will be used across all training methods. 

\begin{table}[!t]
\centering
\caption{Optimised Hyperparameters for Neural Network}
\label{tab:7}
\begin{tabular}{lr}
\hline
Hyperparameter           & Value \\ \hline
Optimiser                & Adam  \\
Learning Rate            & 0.001 \\
Epochs                   & 8     \\
Time Steps               & 1     \\
Training samples         & 4,000,000  \\
Batch Size               & 256   \\
Input Branch Layer Width & 32    \\
Output Layer Width       & 96    \\ \hline
\end{tabular}
\end{table}

As Table \ref{tab:7} shows, the NN trains best with 4M samples. Since the healthy data from each dataset contains over 4M samples, each file from the training set was randomly sliced to reduce the number of samples from each file, totalling the 4M samples required for each dataset when concatenated. This approach ensures that the model will train on a wide variety of data, and subsequently increase the reliability of the experimental results by avoiding optimising for a specific set of files. 

There are various methods of determining the error thresholds of an Autoencoder NN. The most common method is through the use of an unseen validation set. After training the NN, the model is run on a file in the same class as the training set, but which has not been used for training. In this case, a single file containing data for a healthy condition motor is used. The Mean Absolute Error (MAE) for each prediction is then calculated, which provides a baseline for the accuracy of the NN with reconstructing healthy data. When multiple features are present in the dataset, as is the case with the data used in this case study, a threshold is calculated for each feature since the reconstruction performance of the NN model may vary across each feature. The overall threshold can be calculated through different methods, and the method used is determined based on the training performance of the NN as well as the data consistency and behaviour. In the present work, two methods are used: The largest MAE for each feature, or two standard deviations away from the mean MAE of each feature. The reasoning behind using two standard deviations from the mean as a threshold is, assuming a Gaussian distribution of error residual values, two standard deviations from the mean covers 95\% of the data. In the case of an inconsistent or noisy dataset, it is better to use this method since there are more likely to be anomalous MAE values in the validation set, and if the largest MAE value is used, the threshold may be too high to consistently detect abnormalities in the test data. In the case of a consistent dataset with contains minimal noise, the largest MAE value is generally a good threshold to use since an MAE value to exceeds this threshold is more likely to indicate an anomaly as opposed to noisy but healthy behaviour.

\subsubsection{Addressing limitations though experimental design}

A potential limitation of the proposed experimentation is the experimentation on static datasets. By experimenting on static datasets, results achieved in experimental conditions may not generalise effectively to real-world conditions or even other data in the same problem domain. One experiment will address this limitation by recreating the dataset for each of the ten test iterations, shuffling the dataset files for each test iteration randomly in each dataset prior to data formulation. 

Another potential limitation is the availability of an equal amount of health data from each Dataset tested. For the purpose of this experimentation, an equal amount of data from each dataset will be used, however, it cannot be ensured that there is an equal amount of healthy data available outside of experimental settings. Therefore, one experiment will address this issue by modifying the ratio of data used from each dataset to obverse the difference that is made to the anomaly detection results.

\subsection{Experiment 1 - Training methods comparison}
Experiment 1 aims to provide an initial comparison of the various training approaches detailed in Section \ref{Experimental Design}. To ensure experimental validity, the experiment will be repeated 10 times. However, each iteration will not utilise the same training data; instead, it will employ a randomly shuffled subset of data from each file to create a dataset comprising 4M samples. This approach further bolsters the validity of the proposed method and mitigates the risk of misrepresenting its performance by only validating it on a single subset of data.

\begin{figure}[!t]
\centering
\includegraphics[width=\linewidth]{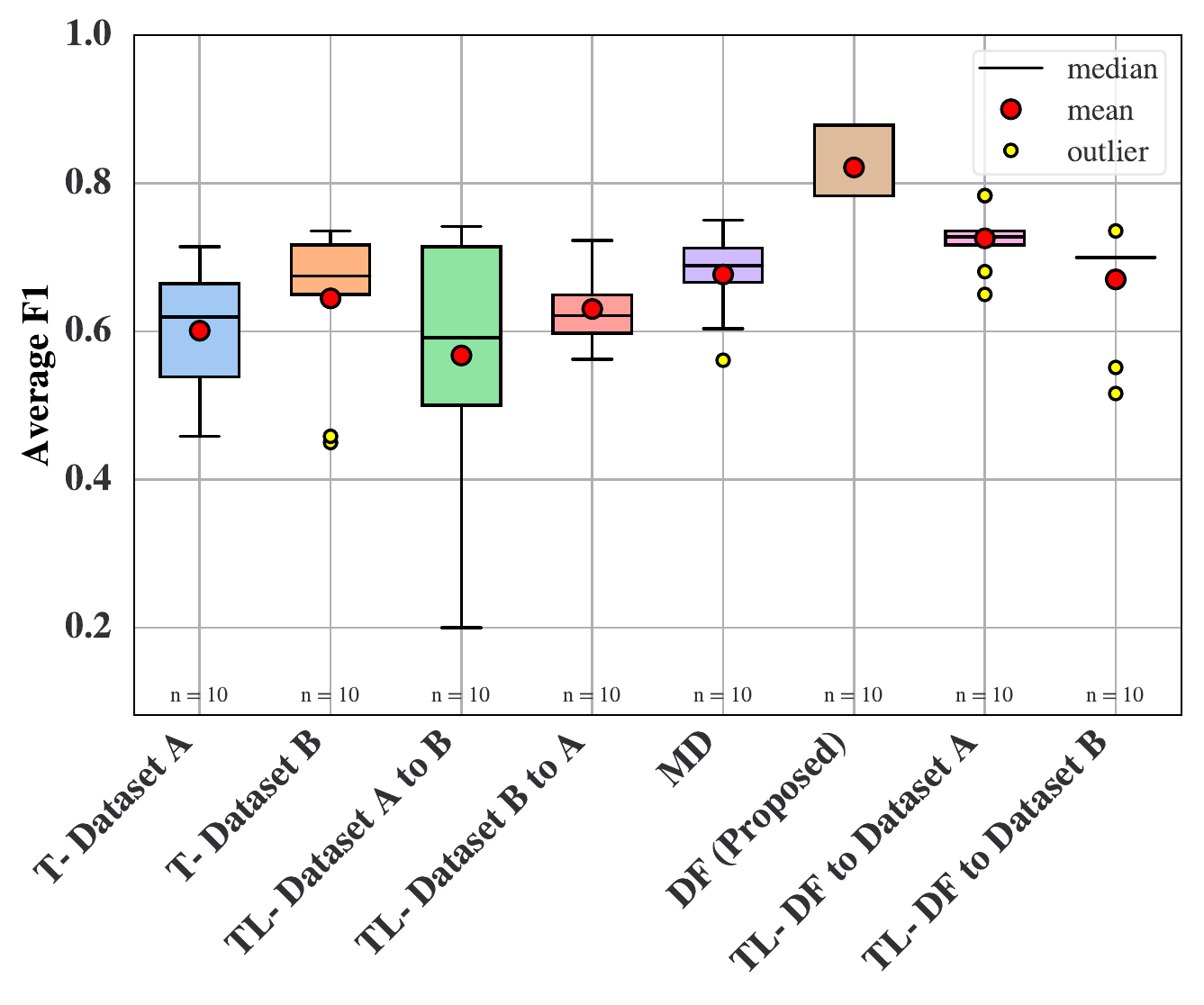}
\caption{Box and whisker plot comparing the results for Experiment 1 - Training methods comparison}
\label{fig:4}
\end{figure}

Table \ref{tab:8} presents the Precision, Recall, and F1 score results for each experiment, as well as the average F1 score across the datasets. Table \ref{tab:9} displays the ANOVA for the Average F1 score results in \ref{tab:6}. Additionally, Figure \ref{fig:4} features a box and whisker plot that visually compares the Average F1 scores from all 10 runs from each training approach, offering a representation of result spread and consistency.

\begin{table*}[!t]
\caption{Experiment 1 results - A comparison of the performance of the anomaly detection model using all training approaches. The experiment key is shown in Table \ref{tab:5}}
\includegraphics[width=\linewidth]{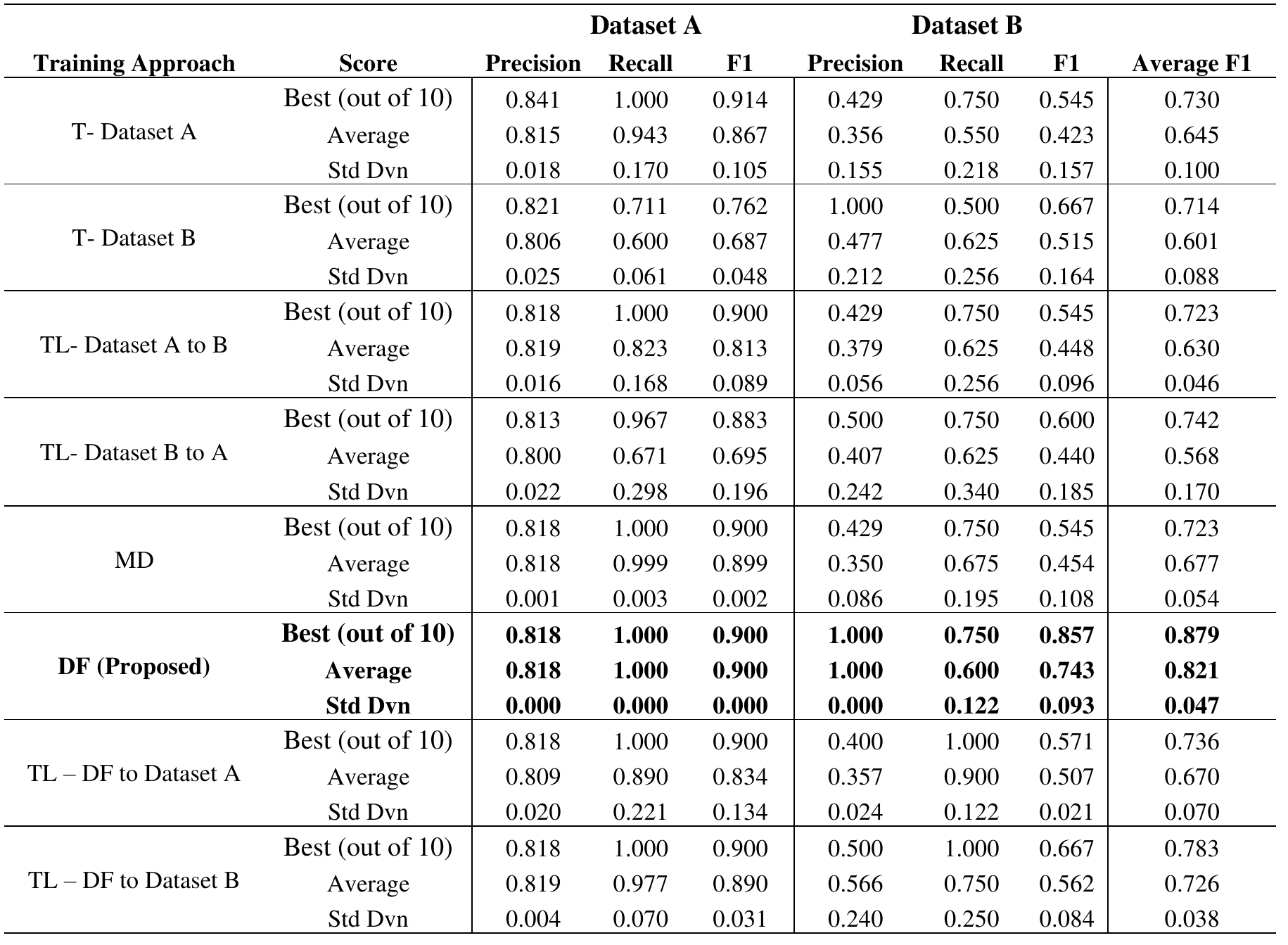}
\label{tab:8}
\end{table*}

\begin{table*}[!t]
\centering
\caption{One-way ANOVA test results for Experiment 1 - Training methods comparison}
\label{tab:9}
\begin{tabular}{lccccc}
\hline
Source         & Degrees of Freedom & Sum of Squares & Mean Square & F-stat & \textit{p}-value  \\ \hline
Between Groups & 7                  & 0.435          & 0.062       & 7.439  & $1.024 \times 10^{-6}$ \\
Within Groups  & 72                 & 0.601          & 0.008       &        &          \\
Total          & 79                 & 1.036          &             &        &          \\ \hline
\end{tabular}
\end{table*}

\subsection{Experiment 2 - Varying data volume}

Experiment 2 aims to observe the effect of reducing the volume of training data on the performance of the anomaly detection model using the different training approaches. Similar to Experiment 1, the experiment will be repeated 10 times, and the dataset will be shuffled to ensure experimental validity. The experiment details for each test are shown in Table \ref{tab:10}. The estimated FLOPs used for each number of samples are calculated using Equation \ref{eq:8} \cite{epoch2022}:

\begin{equation}
    FLOPs = 2 \cdot P \cdot 3 \cdot S \cdot E
\label{eq:8}
\end{equation}
\noindent where $P$ is the number of trainable parameters in the NN, $N$ is number of training samples, and $E$ is the number of epochs. Whilst the complexity of the model can undoubtedly affect the complexity of training, since the same model is used across all experiments, it will not be relevant for this calculation.

\begin{figure*}[!t]
\centering
\includegraphics[width=\linewidth]{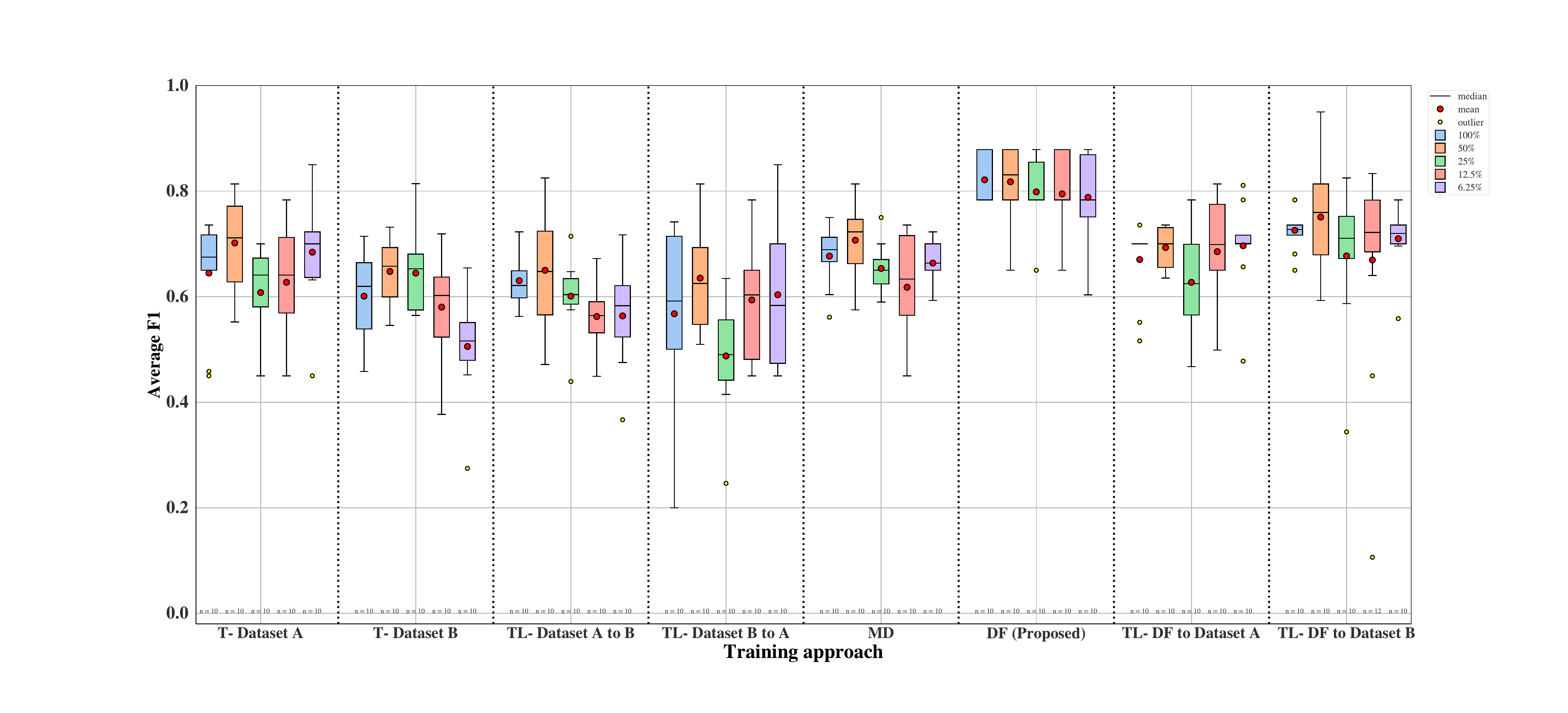}
\caption{Box and whisker plot showcasing the impact of varying data volumes on each training method, grouped by training approach and plotted against average F1 score (higher is better).}
\label{fig:5}
\end{figure*}

The tabulated results for experiment 2 can be found in Table \ref{tab:11}. A summary of the results in the form of a box and whisker plot is illustrated in Figure \ref{fig:5}. The One-way ANOVA table for the Average F1 scores is shown in Table \ref{tab:12}.

\begin{table}[!t]
\caption{Experiment settings for Experiment 2 - Varying data volume, including an estimation of the FLOPs used for training}
\begin{tabular}{crccc}
\hline
\begin{tabular}[c]{@{}c@{}}\% Training\\ data used\end{tabular} & \begin{tabular}[c]{@{}c@{}}Number of\\ Samples\end{tabular} & \begin{tabular}[c]{@{}c@{}}Number of\\ Epochs\end{tabular} & \begin{tabular}[c]{@{}c@{}}NN Trainable\\ Parameters\end{tabular} & \begin{tabular}[c]{@{}c@{}}FLOPs used\\ for training\end{tabular} \\ \hline
100\%                                                       & 4,000,000                                                   & 8                                                          & 25,635                                                            & $4.92 \times 10^{12}$                                     \\
50\%                                                         & 2,000,000                                                   & 8                                                          & 25,635                                                            & $2.46 \times 10^{12}$                                     \\
25\%                                                         & 1,000,000                                                   & 8                                                          & 25,635                                                            & $1.23 \times 10^{11}$                                     \\
12.5\%                                                       & 500,000                                                     & 8                                                          & 25,635                                                            & $6.15 \times 10^{11}$                                     \\
6.25\%                                                       & 250,000                                                     & 8                                                          & 25,635                                                            & $3.08 \times 10^{11}$                                     \\ \hline
\end{tabular}
\label{tab:10}
\end{table}

\begin{table*}
  \caption{Full results of experiment 2 - Varying data volume}
  \label{tab:11}
  \includegraphics[width=\linewidth]{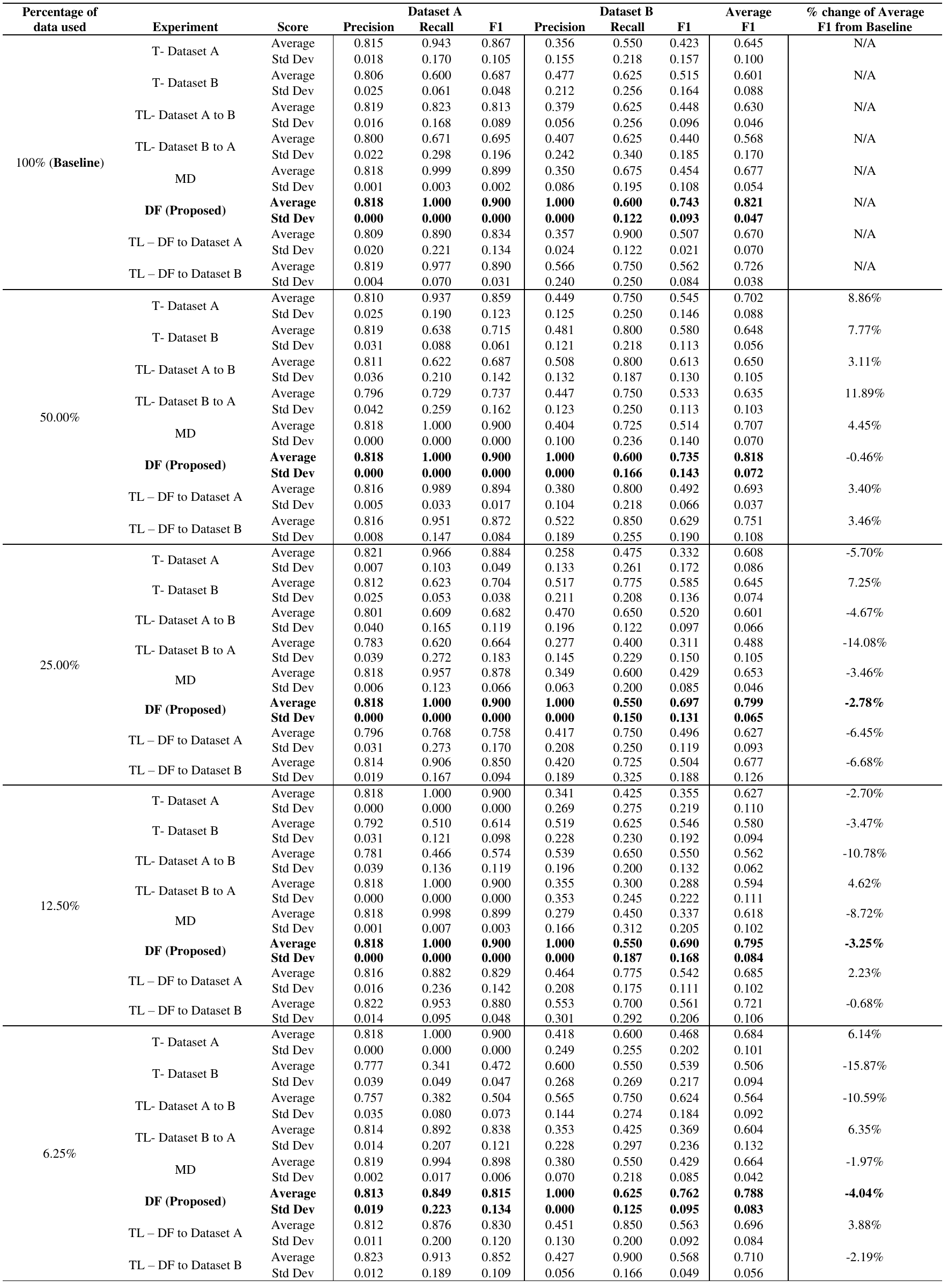}
\end{table*}

\begin{table*}[!t]
\centering
\caption{One-way ANOVA test results for Experiment 2 - Varying data volume}
\label{tab:12}
\begin{tabular}{lccccc}
\hline
Source         & Degrees of Freedom & Sum of Squares & Mean Square & F-stat & \textit{p}-value  \\ \hline
Between Groups & 39                 & 2.389          & 0.061       & 6.341  & $1.110 \times 10^{-16}$ \\
Within Groups  & 362                & 3.497          & 0.010       &        &          \\
Total          & 401                & 5.886          &             &        &          \\ \hline
\end{tabular}
\end{table*}

\subsection{Experiment 3 - Varying dataset ratio}

Experiment 3 aims to assess the performance of each training method with an imbalanced dataset containing a different number of samples from each dataset. The purpose of this experiment is to simulate a real-world environment, where the volume of data available from different sources will not be equal in many cases. Similar to Experiments 1 and 2, the experiment will be repeated 10 times, and the dataset will be shuffled to ensure experimental validity.

Table \ref{tab:13} shows a breakdown of the experimental settings used. In the case of traditional training, the anomaly detector model will be trained on the reduced dataset, similar to experiment 2. However, transfer learning approaches will make use of both datasets.

\begin{figure*}[!t]
\centering
\includegraphics[width=7.3in]{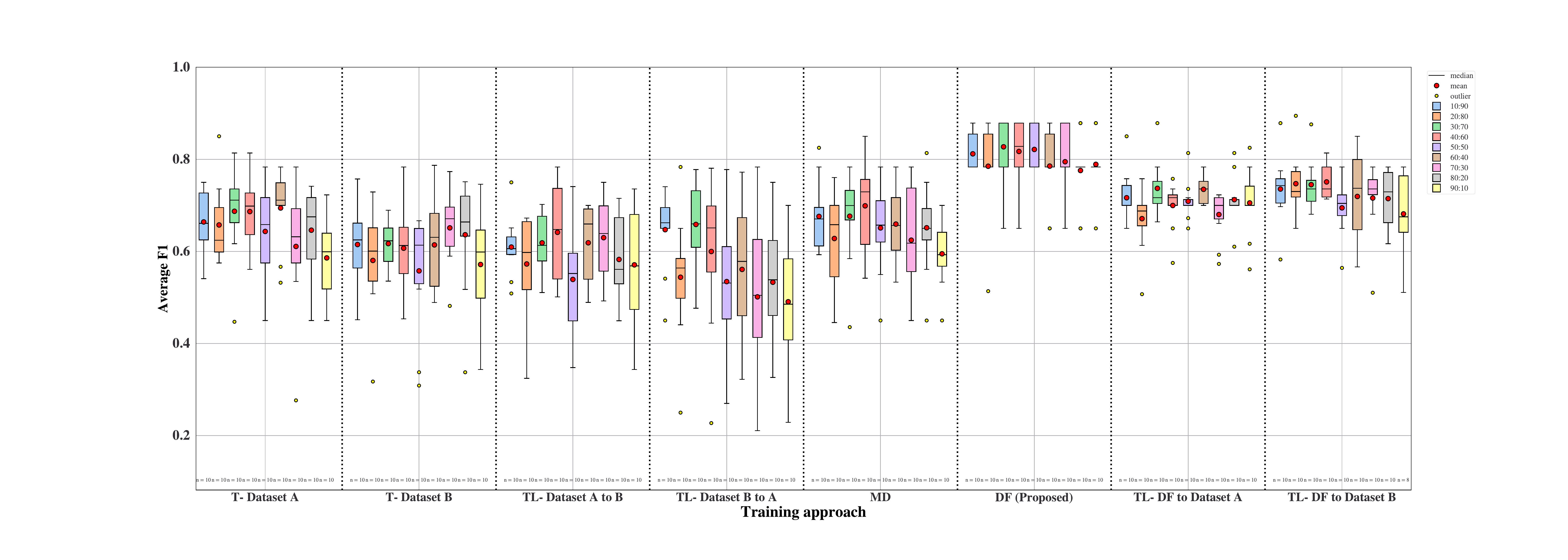}
\caption{Box and whisker plot showcasing the impact of varying dataset ratios on each training method, grouped by training approach and plotted against average F1 score (higher is better).}
\label{fig:6}
\end{figure*}

The tabulated results for experiment 3 can be found in Table \ref{tab:14} for results from 10:90 to 50:50 (Dataset A: Dataset B), and in \ref{tab:15} for results from 60:40 to 90:10. The tabulated results are summarised in a box and whisker plot, shown in Figure \ref{fig:6}. The One-way ANOVA table for the Average F1 scores is shown in Table \ref{tab:16}.

\begin{table}[]
\centering
\caption{Experiment 3 - Varying dataset ratio details}
\label{tab:13}
\begin{tabular}{@{}crr@{}}
\hline
Dataset Ratio (A:B) & Dataset A samples & Dataset B samples \\ \hline
10:90               & 400,000            & 3,600,000           \\
20:80               & 800,000            & 3,200,000           \\
30:70               & 1,200,000           & 2,800,000           \\
60:40               & 1,600,000           & 2,400,000           \\
50:50               & 2,000,000           & 2,000,000           \\
40:60               & 2,400,000           & 1,600,000           \\
30:70               & 2,800,000           & 1,200,000           \\
20:80               & 3,200,000           & 800,000            \\
90:10               & 3,600,000           & 400,000            \\ \hline
\end{tabular}
\end{table}

\begin{table*}
\centering
  \caption{Results for Experiment 3 - Dataset Ratio from 90:10 to 50:50 (Dataset A : Dataset B)}
  \label{tab:14}
  \includegraphics[width=6in]{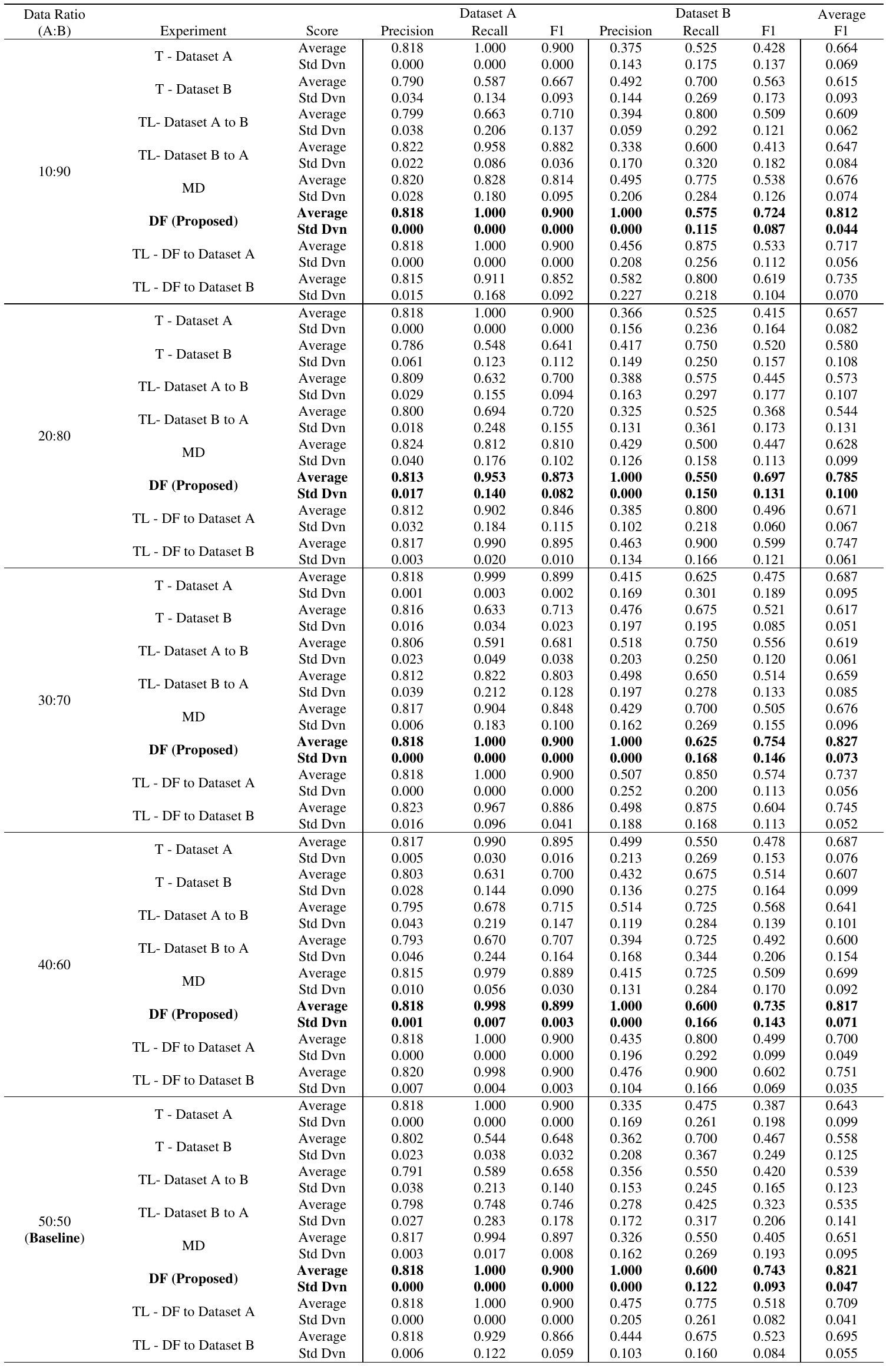}
\end{table*}

\begin{table*}
\centering
  \caption{Results for Experiment 3 - Dataset Ratio from 40:60 to 10:90 (Dataset A : Dataset B)}
  \label{tab:15}
  \includegraphics[width=\linewidth]{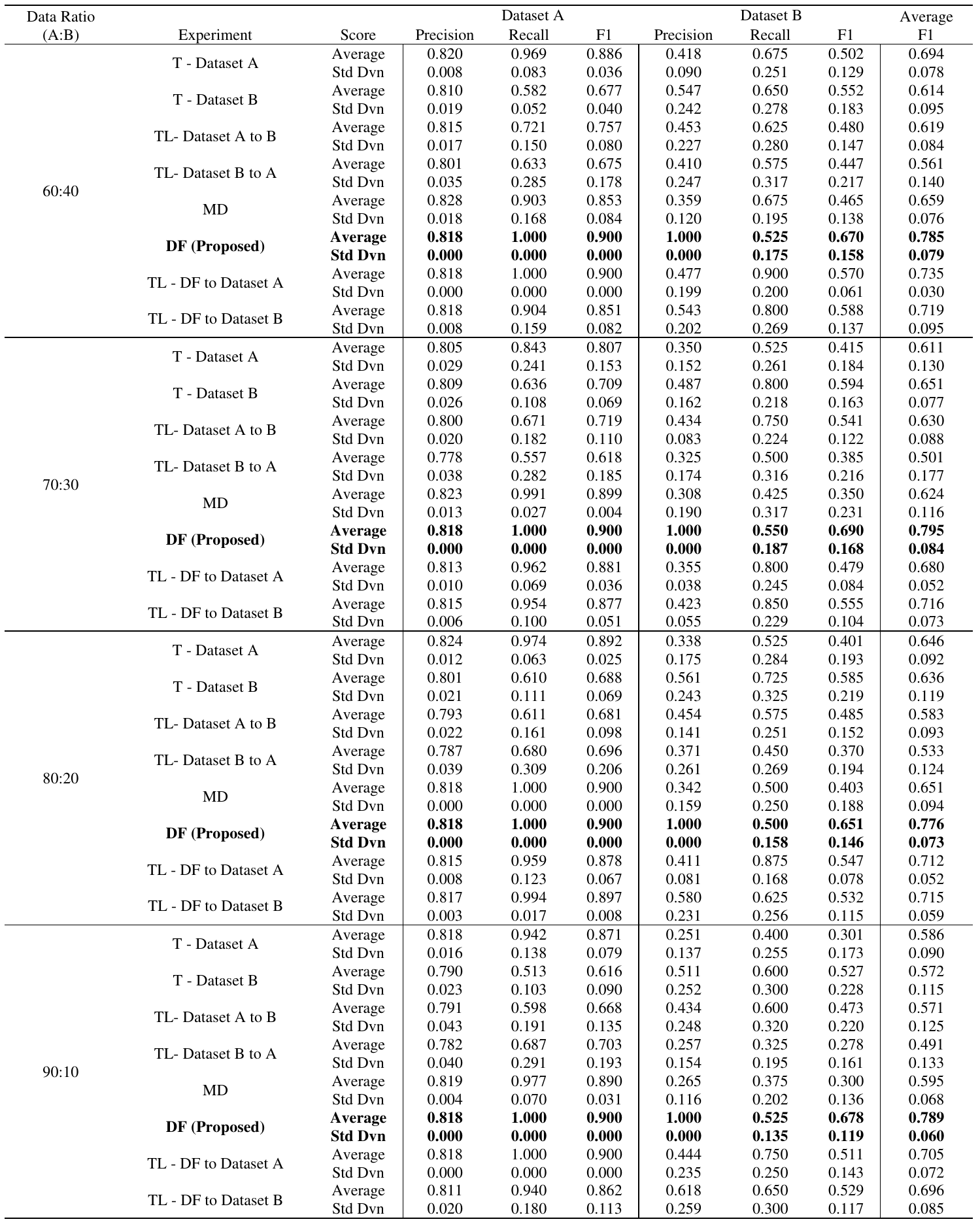}
\end{table*}

\begin{table*}[!t]
\centering
\caption{One-way ANOVA test results for Experiment 3 - Varying dataset ratio}
\label{tab:16}
\begin{tabular}{llllll}
\hline
Source         & Degrees of Freedom & Sum of Squares & Mean Square & F-stat & \textit{p}-value  \\ \hline
Between Groups & 71                 & 4.546          & 0.064       & 6.935  & $1.110 \times 10^{-16}$ \\
Within Groups  & 646                & 5.964          & 0.009       &        &          \\
Total          & 717                & 10.510         &             &        &          \\ \hline
\end{tabular}
\end{table*}

\section{Discussion}

\subsection{Experiment 1 - Training methods comparison Analysis}

Examining the experimental results from Experiment 1, as presented in Table \ref{tab:8}, the DF method is empirically proven to consistently deliver the best performance in terms of F1 score for both Dataset A and Dataset B, as well as the average F1 score across both datasets. With the best Average F1 score of 0.879 and a 10-run average of 0.821, DF surpasses the other methods. These findings suggest that the DF approach effectively captures the salient features in both datasets, resulting in consistently strong performance across the datasets compared to the compared methods. Moreover, the results imply a considerable advantage in fusing the datasets, as the performance on individual datasets significantly exceeds that of models specialised in each respective datasets.

In comparison, the traditional training approach on Dataset A (T-Dataset A) exhibits better performance on Dataset A with a mean F1 score of 0.867 but suffers from poor results on Dataset B with a weaker score of 0.423, consequently lowering the average F1 score and making this model unsuitable for use across homogeneous datasets. The superior performance of the DF method on both Dataset A and Dataset B, along with the average F1 score across both datasets, underscores the algorithm's effectiveness in fulfilling the need for a single neural network adaptable to data from various sources within the same problem domain. This outcome aligns with the algorithm's proposed benefits, which aim to eliminate the need for multiple NNs and reduce data requirements from individual sources.

Furthermore, the results indicate that the traditional training approach, while effective for the dataset it was trained on, falls short in terms of generalisability across homogeneous datasets. In contrast, the DF method not only provides a more robust solution for handling data from multiple sources but also proves to be consistent in performance across multiple runs.

Utilising transfer learning, even when first training with the fused dataset, does not lead to better overall performance, even on the dataset that was trained on in the second phase. However, the preprocessing methods employed in the DF algorithm play a crucial role in enhancing performance, particularly for Dataset B, which required downsampling to meet the algorithm's requirements. This demonstrates the algorithm's adaptability to variations in data characteristics and its potential for handling real-world scenarios where data collection specifications may be inconsistent.

Although the performance across different training approaches on Dataset B is lower than on Dataset A, using transfer learning from the fused dataset to Dataset B yields results that are on par with training solely on Dataset B using the traditional approach. Simultaneously, this approach achieves strong performance on Dataset A. However, the DF algorithm still outperforms all transfer learning approaches in terms of overall performance on both test datasets. 

The box and whisker plot in Figure \ref{fig:4}, illustrating the spread of the results, shows that the DF approach did not produce any outlying results from the 10 runs, indicating high levels of consistency. However, it is also evident that the transfer learning approaches exhibit the lowest deviation in results compared to a single training phase on one dataset, whether using traditional training approaches or the proposed DF method. This highlights the potential benefits of combining transfer learning with the DF approach to achieve even more consistent and reliable performance across datasets. However, the results show that transfer learning may not be the best approach, especially in this problem domain, to achieve the best performance.

The ANOVA Analysis presented in Table \ref{tab:9} demonstrates that the experimental results yield a statistically significant outcome, with a \textit{p}-value of 1e-16. This highlights the relevance of the differences observed among the various training approaches.

\subsection{Experiment 2 - Varying data volume Analysis}

The results of Experiment 2, presented in Table \ref{tab:11} and visualised in Figure \ref{fig:5}, demonstrate that the DF approach significantly outperforms other training approaches across varying volumes of training data. Furthermore, when utilising only 6.25\% of the training data, the model still surpasses other training approaches that use 100\% of the data, achieving an Average F1 score of 0.788. As expected, most approaches experience decreased performance when using less data, with the proposed DF approach following this trend closely. However, it is worth noting that the performance reduction is minimal, with only a 4.04\% drop compared to the baseline results.

Interestingly, not all approaches follow this pattern. For instance, transfer learning from the fused dataset to dataset B performs best when using 50\% of the data, while transfer learning from the fused dataset to dataset A achieves optimal results with only 6.25\% of the data. For these approaches, there does not appear to be a clear advantage to using more data, leading to the conclusion that, for training unsupervised anomaly detectors, using more data is not always beneficial. While this may not hold true for other tasks such as fault classification, those tasks are beyond the scope of the present study and will be explored in future works.

The box and whisker plot in Figure \ref{fig:5} reveals that the performance of the anomaly detection model trained with DF is less consistent and has a larger spread when using less data. However, examining the results from other training approaches shows that this is not always the case. In some instances, such as Transfer Learning from Dataset A to Dataset B, the spread is largest when training with the full dataset. Additionally, as the ANOVA table in Table \ref{tab:12} shows, the \textit{p}-value for this experiment suggests the results are statistically significant.

By outperforming all other models while using only a small fraction of the training data, the DF method addresses the common challenge of not having sufficient training data from each data source. In this experiment, as shown in Table \ref{tab:10}, the estimated FLOPs needed for training the model with 6.25\% of the data dramatically decreased from $4.92 \times 10^{12}$ to $3.08 \times 10^{11}$, representing a 93.7\% reduction in computational power. This significant decrease is especially noteworthy when contrasted with the minor 4.04\% reduction in performance. The DF approach effectively utilises less data, showcasing its potential to contribute to more sustainable and environmentally friendly AI development. Aligned with the principles of Green AI, which emphasise efficiency and reducing the environmental impact of training AI models, the results of Experiment 2 highlight the superior performance of DF. Its crucial implications for real-world applications demonstrate its ability to address the issue of limited training data while promoting more sustainable practices in AI development.

\subsection{Experiment 3 - Varying dataset ratio Analysis}

The results of the experiment, shown in Table \ref{tab:14} and Table \ref{tab:15}, reveal that the DF algorithm outperforms all other training approaches in terms of Average F1 score. The best performance is achieved with a 30:70 ratio (Dataset A: Dataset B), resulting in an Average F1 score of 0.827, closely followed by the baseline experiment (50:50) with an Average F1 of 0.821. In comparison, the next best performing training approach was transfer learning from the fused dataset to Dataset B (TF – DF to Dataset B), which had an Average F1 score of 0.751. The ANOVA in Table \ref{tab:16} confirms that these values are statistically significant.

The box and whisker plot in Figure \ref{fig:6} indicates that the transfer learning approaches involving the fused dataset generally exhibit less spread compared to other methods. Furthermore, the spread of the DF results appears to be the most consistent across different experimental settings. One might hypothesise that the box plot trend would indicate an increase in generalised performance as the balance between samples from each dataset increases. Interestingly, this is not the case.

While the results empirically demonstrate that an imbalance in datasets mostly does not affect the stability of the results for DF, a clear trend is observed, wherein the average F1 score decreases as more of Dataset A is used for training. This trend is more evident in some of the other approaches, such as T-Dataset A, TL-Dataset B to A, and the mixed dataset (MD). For instance, training with the Mixed Dataset at a 40:60 ratio yields an F1 score of 0.889 on Dataset A, 0.509 on Dataset B, and an overall average of 0.699. Conversely, training with a 90:10 ratio results in an F1 score of 0.890 on Dataset A, 0.300 on Dataset B, and an overall average of 0.595. These results show that the performance on Dataset A can be maintained with less data while improving the performance on Dataset B with more data.

The observed trend leads to an intriguing conclusion: there is a higher benefit to training with Dataset B compared to Dataset A to achieve a more generalised anomaly detection performance. Although both Dataset A and Dataset B represent healthy data, Dataset B exhibits a higher level of noise, which can be clearly seen in Figure \ref{fig:2}h, the raw FFT of the signal. Additionally, there is a slightly higher level of spread on the PCA shown in Figure \ref{fig:3}. These observations suggest that Dataset B is a more difficult dataset to learn and requires more training data compared to Dataset A. This also implies that a larger variety of healthy data in the training dataset contributes to an overall better anomaly detection performance.

\subsection{Further Remarks}

In the context of Transfer Learning, it is important to note that better results on the dataset trained on in the second phase are not guaranteed. The results of the experiments demonstrate that swapping the transfer learning training phases produces different outcomes, which indicates that there is no clear approach to determine which dataset should be used in each phase without conducting additional testing.

The conclusions drawn from Experiment 3 suggest that there is a greater benefit to training on a higher number of samples from Dataset B, as opposed to Dataset A, in order to achieve a more generalised performance. However, identifying this preference presents a challenge, as a similar experiment must be conducted to reach this determination. Despite this issue, the DF approach offers a solution by providing a more stable performance across multiple experimental settings. This advantage becomes particularly significant in practical applications where time and resource constraints may render extensive experimentation unfeasible. By addressing these concerns and offering more consistent results, DF shows potential as an effective method for training unsupervised anomaly detection models, particularly in situations where the optimal dataset and training phase cannot be easily determined. The robust performance of the DF approach, combined with its ability to accommodate various experimental settings, positions it as a valuable tool for practical use cases. It is currently unclear whether the findings from these experiments will hold true for other tasks, such as fault classification. Addressing this question is beyond the scope of this paper, but it presents an intriguing avenue for future research.

\section{Conclusion}

This paper presents a time-series dataset composition approach called the DF algorithm, designed to address challenges in achieving generalised anomaly detection performance across multiple homogeneous data sources. The proposed algorithm was validated using a case study involving motor current signals, demonstrating that the fused dataset retains salient features from both source datasets while clustering in the middle of both datasets when PCA is applied. The algorithm was then tested on an anomaly detection task and compared to conventional training approaches, with empirical results showing that the DF algorithm significantly outperforms other methods in terms of average performance across both datasets.

Additionally, further experiments were conducted to assess the performance of the proposed approach under non-ideal conditions. Experimental results indicate that the DF approach remains superior even when reducing the number of data samples, with only a 4.04\% reduction in performance despite using only 6.25\% of the training data, resulting in a 93.7\% reduction in computational power required for training. When evaluating the model's performance with imbalanced numbers of samples from each dataset, the proposed approach proved stable across different sample ratios. These findings highlight significant benefits in the context of Green AI, which emphasises sustainable AI model development, as well as practical feasibility due to the algorithm's resilience under non-ideal conditions.

Several research directions could be explored based on the present study. For instance, future works might investigate the applicability of the algorithm for training classifier models or examine whether using the frequency domain representation of the fused dataset could enhance performance. Furthermore, it would be worthwhile to explore the relationship between dataset "complexity" and the "usefulness" of data from each fused dataset, enabling the development of a more systematic approach to dataset fusion.

\section*{Acknowledgments}
The authors would like to express their sincere gratitude to Voltvision for their substantial support of this research. Their sponsorship has significantly contributed to the progress and achievements of this work.

\bibliographystyle{IEEEtran}
\bibliography{references}

\section*{Biography Section}

\vspace{-8 mm}

\begin{IEEEbiography}[{\includegraphics[width=1in,height=1.25in,clip,keepaspectratio]{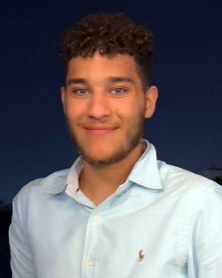}}]{Ayman Elhalwagy}
received the BEng (Hons) degree in electronic and computer Engineering from Brunel University London, Uxbridge in 2021 and is pursuing the Ph.D. degree in Electronic and Electrical Engineering with Brunel University London, Uxbridge.
His research interests include Applied Machine Learning, Fault Detection and Classification, and intelligent systems.
\end{IEEEbiography}

\vspace{-8 mm}

\begin{IEEEbiography}[{\includegraphics[width=1in,height=1.25in,clip,keepaspectratio]{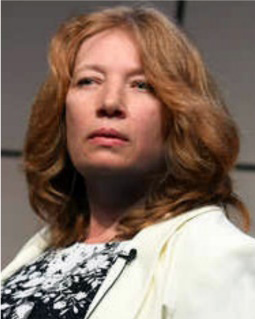}}]{Tatiana Kalganova}
received the B.Sc. (Hons.) and Ph.D. degrees. She is currently a Professor in intelligent systems and the Electronic and Computer Engineering Postgraduate Research Director with Brunel University London, Uxbridge, U.K. She has over 30 years of experience in the design and implementation of applied intelligent systems. 
\end{IEEEbiography}

\vfill
\end{document}